\documentclass[11pt]{article}

\usepackage[preprint]{acl}

\usepackage{times}          
\usepackage{latexsym}
\usepackage[T1]{fontenc}    
\usepackage[utf8]{inputenc} 
\usepackage{inconsolata}    
\usepackage{hyperref}       
\usepackage{url}            
\usepackage{booktabs}       
\usepackage{amsmath}        
\usepackage{amsfonts}       
\usepackage{amssymb}        
\usepackage{nicefrac}       
\usepackage{microtype}      
\usepackage{xcolor}         
\usepackage{graphicx}
\usepackage{multirow}
\usepackage{algorithm}      
\usepackage{algpseudocode}  
\usepackage{fvextra}        

\graphicspath{{figures/}}

\newif\ifshowcomments
\showcommentsfalse

\usepackage[normalem]{ulem}  
\ifshowcomments
  \newcommand{\huhan}[1]{\textcolor{pink}{\textbf{[HH: #1]}}}
  \newcommand{\glu}[1]{\textcolor{teal}{\textbf{[GLU: #1]}}}
  \newcommand{\sfei}[1]{\textcolor{blue}{\textbf{[SF: #1]}}}
  \newcommand{\ysun}[1]{\textcolor{orange}{\textbf{[ysun1: #1]}}}
  \newcommand{\shlong}[1]{\textcolor{purple}{\textbf{[shlong: #1]}}}
  \newcommand{\kding}[1]{\textcolor{olive}{\textbf{[KD: #1]}}}
  \newcommand{\fengyi}[1]{\textcolor{red}{\textbf{[fengyi: #1]}}}
  \newcommand{\resolved}[1]{\textcolor{gray}{\sout{[RESOLVED] #1}}}
\else
  \newcommand{\huhan}[1]{}
  \newcommand{\glu}[1]{}
  \newcommand{\sfei}[1]{}
  \newcommand{\ysun}[1]{}
  \newcommand{\shlong}[1]{}
  \newcommand{\kding}[1]{}
  \newcommand{\fengyi}[1]{}
  \newcommand{\resolved}[1]{}
\fi

\title{SPEAR: Code-Augmented Agentic Prompt Optimization}

\author{%
  \parbox[t]{0.92\textwidth}{\centering
    Mengyin Lu, Cong Feng, Huimin Han, Guangming Lu, Yu Sun, \\
    Xiaonan Ding, Shihui Long, Fengyi Li, Tanvi Motwani
  } \\
  LinkedIn Corporation \\
  \texttt{\{melu, cofeng, huhan, glu, ysun1, kding, shlong, fenli, tmotwani\}@linkedin.com}
}

\begin{document}

\maketitle
\resolved{\sfei{Maybe change the name to something like: ``SPEAR: Code-as-Action Agentic Prompt Optimization with Guard-Metric Rollback''?}}\sfei{Adopted SPEAR as the lead identifier in a shorter form: ``SPEAR: Code-Augmented Agentic Prompt Optimization.'' Dropped the ``Guard-Metric Rollback'' suffix because the guard-metric was downgraded from a numbered contribution to method-section glue.}

\begin{abstract}
Automatic prompt engineering (APE) rewrites prompts to improve
downstream task performance, but existing APE loops treat the
optimizer itself as a fixed pipeline. We port the code-as-action
paradigm of CodeAct~\citep{wang2024codeact} to APE and propose
\emph{SPEAR} (Sandboxed Prompt Engineer with Active Rollback), a
free-form agentic optimizer with four tools --
\texttt{evaluate}, \texttt{python}, \texttt{set\_prompt},
\texttt{finish} -- that decides autonomously how and when to use
them. The distinctive tool is the Python sandbox: the optimizer
writes and executes arbitrary Python on the current evaluation
DataFrame, performing structural error analysis (confusion
matrices, error clustering, per-group metrics) the agent itself
authors. Two guardrails keep the long-horizon agent from regressing
on the validation metric: auto-rollback to the best validation
prompt seen, and an optional guard-metric floor. We evaluate on three
industrial LLM-as-judge suites (13 judge tasks across
recruiter-intake, conversational-memory, and query-refinement
systems) plus seven BBH tasks and GSM8K. In a like-for-like
comparison (same optimizer, matched runs) SPEAR leads every
industrial task it is compared on ($\kappa$ 0.857 vs 0.359 on
tool-selection; F1-macro 0.815 vs 0.763 on filter-relevance;
$\kappa$ 0.483 vs 0.260 on the hardest extraction dimension). On
BBH-7 SPEAR averages 0.938 accuracy vs GEPA 0.628 and TextGrad
0.484. On the public Banking77 benchmark, from a shared seed prompt,
only SPEAR substantially improves ($\kappa$ 0.927 vs GEPA 0.816 and
instruction-only MIPRO 0.753), leakage-free evidence for the
confusion-analysis mechanism. Ablations show the Python tool is the largest single lever
on complex judge tasks ($\Delta\approx{+}0.79\kappa$ on the 5-class
tool-selection judge, $\Delta\approx{+}0.35\kappa$ on the hardest
extraction dimension when removed); its irreplaceable contribution is
class-pair confusion aggregation that a long-context LLM cannot
extract reliably from the raw eval DataFrame.
\end{abstract}
\resolved{\glu{The abstract could be crisper.}}
\resolved{\glu{Remove "(OPRO, EvoPrompt, TextGrad, GEPA, DSPy/MIPRO, PromptAgent,
ProTeGi, Trace/OPTO)"?}}
\resolved{\glu{Remove "In a like-for-like
comparison where all methods use GPT-5.4 as the optimizer,"?}}
\resolved{\glu{Remove "a large share of this gap is ... tools underperforms the free-form agent." ?}}\glu{All four applied.
Abstract is now $\sim$160 words and contains only: problem framing,
SPEAR sketch, two guardrails, dataset list, headline numbers,
ablation lever.}

\resolved{\huhan{Abstract is long (250+ words). NeurIPS typically expects $\sim$150--200 words. Consider cutting the BBH honest-disclosure sentence and the held-out/seed-variance sentence --- those belong in the body. Also: guard-metric is mentioned once but never quantified here; either add a number or drop it from the abstract.}}

\resolved{\glu{Should we rephrase LIHA/CMA to some more external-facing terms?}}

\resolved{\ysun{GSM8K seed accuracy is already 0.965 (Tab.~\ref{tab:bbh}); GEPA lands at 0.960 and TextGrad at 0.956 --- both \emph{below} seed. ``All three methods tie at $\approx$0.96'' is technically true but obscures that there is zero headroom above the seed and two baselines actually regressed. Recommend reframing as ``no method moved the GSM8K metric'' --- this actually \emph{strengthens} the format-vs-reasoning argument rather than reading as a wash.}}

\section{Introduction}
\label{sec:intro}

Automatic prompt engineering (APE) rewrites a seed prompt to
maximise a metric on a labeled dataset. Existing
systems~\citep{yang2024opro, guo2024evoprompt,
yuksekgonul2024textgrad, agrawal2025gepa} all share one structural
trait: the optimizer is a \emph{fixed pipeline}. Its error signal
-- scalar loss, Pareto front, or single-row textual critique -- is
chosen at design time, and the analysis the optimizer performs is
whatever the pipeline pre-bakes. When errors cluster by structure
(a confused class pair, a label-rule contradiction visible only
when grouping by a feature), a fixed pipeline cannot expose that
structure to the optimizer.

\paragraph{Our proposal.}
We replace the fixed APE pipeline with a free-form agent equipped
with a Python sandbox, transplanting the \emph{code-as-action}
paradigm of CodeAct~\citep{wang2024codeact} into the APE loop.
\textbf{SPEAR} has four tools -- \texttt{evaluate} (score the
current prompt and return per-row predictions), \texttt{python}
(run arbitrary code over that DataFrame), \texttt{set\_prompt}
(rewrite), \texttt{finish} -- and is the first APE system in
which the optimizer \emph{actively authors} analysis code over the
evaluation DataFrame (positioning in
Appendix~\ref{sec:app-novelty-table}).\resolved{\huhan{``to our knowledge'' 是 reviewer red flag for under-surveyed related work。直接说 ``and is the first APE system in which...'' 就行 --- Tab.~\ref{tab:novelty} 已经系统 survey 过了，有底气直接 claim。}} A concrete payoff: on a 5-class
CMA judge whose seed prompt yields $\kappa$=0.20 because it
collapses minorities into the majority, SPEAR's first
\texttt{python} block builds the EXPECTED$\times$PARSED confusion
matrix, targets the most-confused class pair, and four such
rewrites take $\kappa$ from 0.20 to 0.95+. Two minimal guardrails
-- metric-based \emph{auto-rollback} on regression and an optional
\emph{guard-metric floor} -- keep the long-horizon agent from
regressing on the validation metric (auto-rollback returns the best
validation prompt seen).

\paragraph{Contributions.}
(1) \textbf{Code-as-action applied to APE.} The first APE system
in which the optimizer actively authors analysis code over the
evaluation DataFrame; the closest prior work,
OPTO/Trace~\citep{cheng2024opto}, treats execution traces as
\emph{passive} feedback.
(2) \textbf{Free-form agent vs.\ fixed pipeline.} An ablation
replacing the agent with a rigid same-tools cycle loses 26\,pp on
BBH \texttt{logical\_deduction\_5obj} and 0.27$\kappa$ on Hiring Assistant \emph{job location}: orchestration autonomy carries a
substantial share of the lift.
(3) \textbf{Industrial case studies of what code-as-action
discovers in APE.} Two studies
(\S\ref{sec:case-study}) expose qualitatively
distinct rewrites the affordance enables -- a
confusion-matrix-driven category-definition rewrite on a
multi-class judge, and a label-rule-contradiction discovery on a
binary judge.

\section{Related work}
\label{sec:related}

Existing APE systems
-- LLM-in-loop search~\citep{zhou2023ape, yang2024opro},
evolutionary methods~\citep{fernando2023promptbreeder,
guo2024evoprompt, zehle2025capo, agrawal2025gepa}, surrogate/beam and
tree search~\citep{opsahl2024mipro, pryzant2023protegi,
wang2023promptagent}, symbolic and joint
search~\citep{schnabel2024sammo, agarwal2024promptwizard},
self-supervised~\citep{xiang2025spo}, memory-augmented~\citep{liang2026memapo, gangireddi2026revere},
gated refinement/compression~\citep{shi2025grace,
xu-etal-2025-procut, jiang-etal-2024-longllmlingua}, and textual-gradient
methods~\citep{yuksekgonul2024textgrad}
-- all use a fixed optimizer loop; the optimizer never authors
analysis code over per-row predictions. The closest adjacent work,
OPTO/Trace~\citep{cheng2024opto}, treats execution traces as
\emph{passive} feedback. A concurrent wave of judge-prompt and
multi-agent APE
systems~\citep{zhou2024zepo, wen2025hpss, li2025apeensemble,
zhang2025mars, zhou2025mass, zhang2025p3, long2025vista,
liu2026skillforge, liu2026evox, rishav2026contraprompt,
liu2026reflectiondark, singhal2026prefpo,
breen2026buildjudgeoptimize} similarly lacks an active
code-execution affordance. Among these, Build/Judge/Optimize
\citep{breen2026buildjudgeoptimize} is closest to our industrial
setting -- it calibrates an LLM-as-judge for a production
multi-agent shopping assistant -- but optimizes via
natural-language reflection on multi-turn trajectories rather
than over-DataFrame code execution. SPEAR ports the
\emph{code-as-action} paradigm of CodeAct~\citep{wang2024codeact}
into APE (lineage: ReAct~\citep{yao2023react},
Reflexion~\citep{shinn2023reflexion},
ADAS~\citep{hu2024adas}); auto-rollback is in the family of
GRACE~\citep{shi2025grace}. We benchmark against TextGrad and GEPA
(both open-source, run unchanged); a full positioning table is in
Appendix~\ref{sec:app-novelty-table}. DSPy/MIPRO~\citep{opsahl2024mipro} jointly searches instructions
\emph{and} few-shot demonstrations, but our
LLM-as-judge prompts cannot accommodate demonstrations (each row is
the held-out scoring target, so no row can serve as an in-context
demo without label leakage), reducing MIPRO to
instruction-only there -- a strict subset of what TextGrad and GEPA
already optimize. We do benchmark instruction-only MIPRO on the
public Banking77 task (\S\ref{sec:public}).
OPRO and EvoPrompt BBH numbers
are cited from the published papers
(Appendix~\ref{sec:app-published}).

\section{Method}
\label{sec:method}

\begin{figure}[t]
  \centering
  \includegraphics[width=0.95\linewidth]{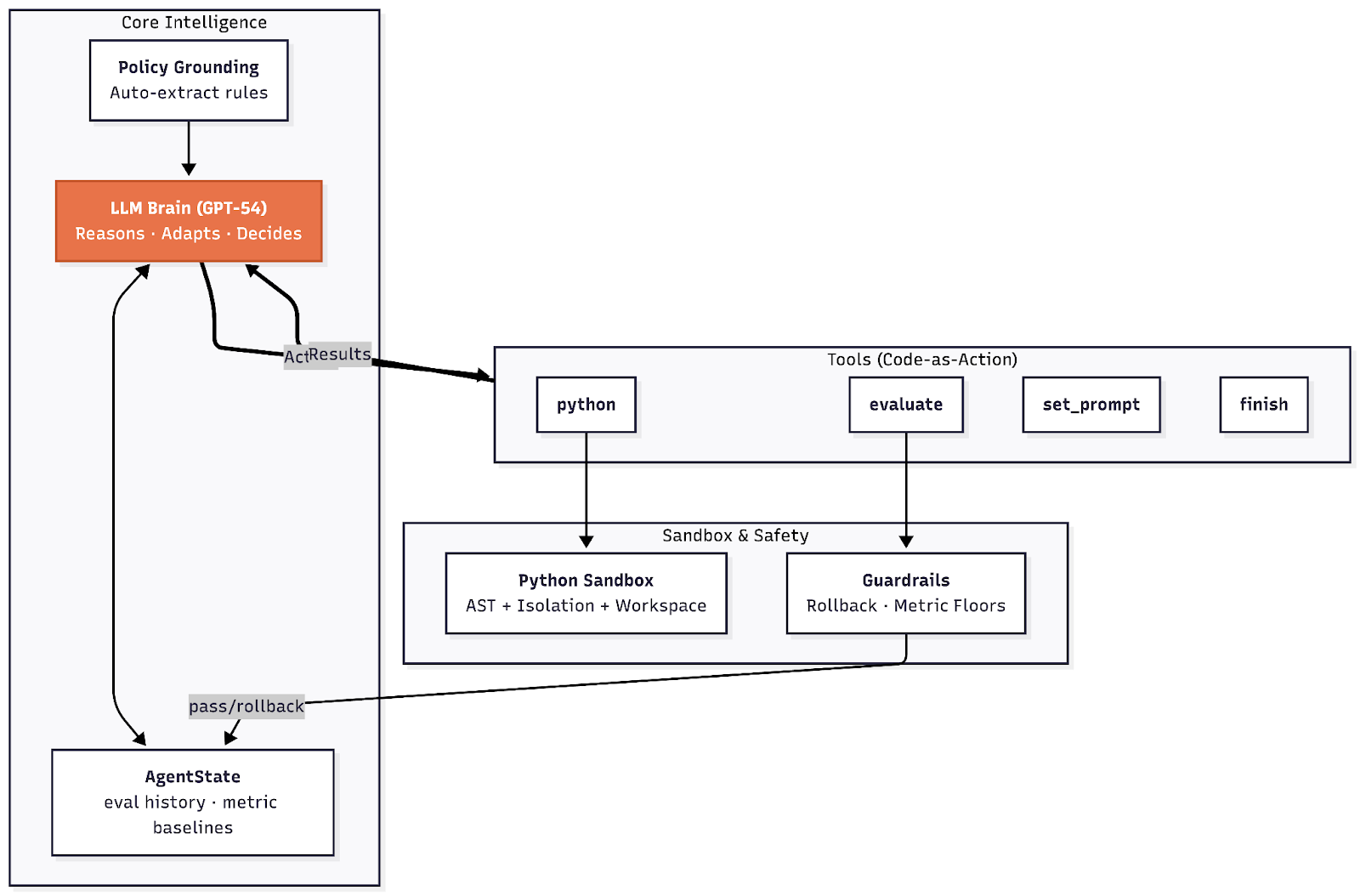}
  \caption{SPEAR architecture. A GPT-5.4 \emph{agent brain} is grounded
  by an optional policy document and chooses one of four tools per
  step: \texttt{python} (executes arbitrary code in an AST-restricted
  sandbox with the current eval DataFrame in scope), \texttt{evaluate}
  (scores the current prompt on train or valid via a task-model
  inference pass), \texttt{set\_prompt} (replaces the system prompt),
  and \texttt{finish} (terminates). Tool outputs (\texttt{Results})
  flow back to the brain and update an \texttt{AgentState} that tracks
  eval history and best-metric baselines. Two guardrails sit between
  tool calls and the state: auto-rollback reverts \texttt{set\_prompt}
  mutations when the primary metric regresses below its best-seen
  value, and a guard-metric floor rejects rewrites that would sink a
  user-nominated secondary metric. No component requires a fixed
  evaluate--analyze--rewrite ordering; the agent decides.}
  \label{fig:architecture}
\end{figure}

\resolved{\glu{Add a problem definition section?}}
\resolved{\glu{Maybe describe in a more formal way like seeks a prompt $P^*$ such that $P^* = \arg\max_P \mathbb{E}[g(f(P(x)), y)]$ ... [original suggestion incorporated into the Problem definition subsection below]}}

\subsection{Four tools, no pipeline}
\label{sec:problem-def}

The optimizer seeks
$P^{*}=\arg\max_{P\in\mathcal{P}} \frac{1}{N}\sum_i
g(\pi(f(P,x_i)),y_i)$ on a labeled dataset
$\{(x_i,y_i)\}_{i=1}^N$, where $f$ is the fixed task model, $\pi$
parses the structured prediction from the generated text, and
$g(\hat{y},y)$ is the primary metric (e.g.\ Cohen's $\kappa$,
macro-F1), subject to a budget of full-dataset
evaluations $\#\text{evals}\le B$ and an optional guard-metric
floor $g_2\ge\tau$ (\S\ref{sec:guardrails}). The dataset is split
into train (analysis), valid (selection), and test (held out);
formal expansion in Appendix~\ref{sec:app-problem-def}.
SPEAR's agent (Figure~\ref{fig:architecture}) runs a plain
\texttt{while} loop and selects one of four tools per step (full
pseudo-code in Appendix~\ref{sec:app-algorithm}):
\textbf{evaluate(split, row\_indices=None)} runs the prompt on
\texttt{train}/\texttt{valid} (optionally a subset) and returns
per-row predictions; one full call consumes a budget unit.
\textbf{python(code)} executes \texttt{code} in a sandbox with
pandas/numpy, the eval DataFrame, the current prompt
(read-only), and an \texttt{llm()} helper (default GPT-4o, $T$=0.2).
\textbf{set\_prompt(new)} replaces the system prompt;
\textbf{finish(summary)} terminates. Defaults:
\texttt{max\_eval\_calls}=20, \texttt{max\_steps}=5$\times$
\texttt{max\_eval\_calls}; the initial \texttt{evaluate} is forced.
Valid is exposed as a \emph{scoring}, not \emph{analysis}, surface:
the sandbox sees aggregate label distributions on valid but never
row content (redaction details and the brain prompt in
Appendices~\ref{sec:app-sandbox-detail},
\ref{sec:app-brain-prompt}).

\resolved{\huhan{The \texttt{llm(system, user)} helper inside the python sandbox is interesting but underspecified. Which model does it call? Does it count against the eval budget? If the agent can call an LLM inside Python, that's a significant capability difference vs baselines that isn't captured in the cost analysis. Clarify or add to the cost accounting in E9.}}

\resolved{\ysun{Distinct from huhan's \texttt{llm()} concern: \texttt{evaluate(split, row\_indices=None)} lets the agent score on a non-representative subset of rows. Two questions: (a) does auto-rollback (\S\ref{sec:guardrails}) compare against the \emph{full-split} best-seen $\kappa$, or against whatever subset was last evaluated? (b) is the headline metric in Tab.~\ref{tab:industrial}/\ref{tab:model-ablation} always recomputed on the full split? If subset metrics can drive rollback or get reported as headline, the agent has an exploit surface for inadvertent metric inflation, distinct from the best-of-$N$ inflation already discussed. Please specify in \S\ref{sec:guardrails}.}}

\resolved{\shlong{Distinct from huhan's \texttt{llm()} concern and ysun1's \texttt{row\_indices} concern (now resolved by the subset-eval guarantee in \S\ref{sec:guardrails}): this targets \emph{read-side} split-level exposure, which the new guarantee does \emph{not} address. The python sandbox lists \texttt{valid\_eval\_df} and raw \texttt{valid\_df} alongside their \texttt{train\_*} counterparts. After a single budgeted \texttt{evaluate(split=`valid')} call (the only kind that updates \texttt{best\_valid\_metric}), the agent may freely \texttt{groupby}, cluster, or otherwise re-aggregate the valid per-row predictions inside the \texttt{python} sandbox at zero further cost, and target subsequent \texttt{set\_prompt} edits at observed valid failures; raw \texttt{valid\_df} is also visible every step regardless of evals. This is a structural asymmetry vs.\ GEPA/TextGrad, whose backward/reflection passes consume only train batches, and is consistent with the empirical signature in Tab.~\ref{tab:heldout} (employment\_type SPEAR dev=0.820 vs.\ test=0.377, gap +0.44). Recommend: (i) state explicitly in \S\ref{sec:method} whether the agent is permitted to read \texttt{valid\_eval\_df} / \texttt{valid\_df} during optimization; (ii) if yes, add an ablation that hides \texttt{valid\_*} from the python sandbox to bound the read-side leakage.}}

\resolved{\shlong{The agent's controlling system prompt --- the ``brain prompt'' that determines when to call \texttt{python} vs.\ \texttt{set\_prompt}, what code style to write, and how to phrase rewrites --- is mentioned here (``encouraged but not required to use the Python tool before \texttt{set\_prompt}'') but never shown anywhere in the paper, including the appendices. Because A1/A4/A5 (\S\ref{sec:ablation}) are all mediated by this prompt --- A4 in particular compares ``free-form agent'' vs.\ ``fixed pipeline built from the \emph{same} tools,'' where the only thing distinguishing the two is the brain prompt --- A4 is unfalsifiable without it. If the brain prompt prescribes, e.g., ``always start with a confusion matrix,'' part of A1's measured ``python-tool effect'' is that instruction's effect, not the tool's. Recommend including the verbatim brain prompt (system prompt plus any per-step status template) in an appendix; this is the minimum required for the ablation chain to be reproducible.}}

\resolved{\shlong{The brain prompt's mandatory ``Use \texttt{llm()} to draft prompts'' (\S\ref{sec:app-brain-prompt}, step 3) calls \texttt{llm()} with \texttt{max\_tokens=16000}, overriding the 4000 default declared above and firing once per \texttt{set\_prompt}. Confirm \S\ref{sec:cost} accounts for this 16k-per-rewrite agent-LM cost.}}

\subsection{Guardrails}
\label{sec:guardrails}

\textbf{Auto-rollback} reverts the prompt when a full-split
\texttt{evaluate} drops the primary metric below its best-seen
value; sampled evaluations do not update best-seen state, so the
agent cannot inflate the headline by selectively evaluating easy
rows. \textbf{Guard metric} is opt-in: a second metric with a floor
$\tau$ that rewrites must not sink below (used on Facet with
\texttt{precision}, $\tau\approx 0.80$ auto-initialised from the
seed prompt; the other 12 dimensions do not need it because
$\kappa$ already penalises majority collapse). Algorithm in
Appendix~\ref{sec:app-algorithm}.
\resolved{\huhan{State the actual Facet guard-metric threshold $\tau$. ``Set automatically from the seed prompt's training-set precision'' is not reproducible without the number. Also: Algorithm~\ref{alg:spear} lines 13--16 show that when $\kappa$ improves but $g_2 < \tau$, the prompt is \emph{neither rolled back nor saved as best} --- it stays as current. On the next noisy eval, the guard-violating prompt could get accepted if $g_2$ coincidentally passes. Is this intended? If so, a sentence clarifying the ``soft reject'' semantics would help; if not, the algorithm should roll back on guard-metric failure too.}}\huhan{Resolved: Algorithm~\ref{alg:spear} fixed -- guard violation now rolls back same as primary-metric drop (the algorithm previously omitted the guard-rollback branch; the actual code rolls back). $\tau$ is auto-initialised from the init-eval guard metric; on Facet this is $\tau\approx 0.80$ (precision).}

\resolved{\huhan{Which tasks actually use the guard metric? Is it only Facet? If so, the guard-metric contribution claim is supported by exactly one task. List all (task, guard-metric, floor-value) triples so reviewers can evaluate the scope.}}

\subsection{Sandbox and safety}
\label{sec:sandbox}

The \texttt{python} tool runs in an AST-restricted sandbox
(module whitelist, no network/shell, 60\,s wall-clock + 512\,MB
RSS cap, read-only prompt) under a cooperative threat model
(trusted optimizer LLM, internal data); sandbox details in
Appendix~\ref{sec:app-sandbox-detail}, eval-tool reliability
instrumentation in Appendix~\ref{sec:app-eval-instrumentation}.

\resolved{\huhan{``200+ logged SPEAR runs'' --- this is weak evidence for safety. Absence of escape attempts under a cooperative model is not a security guarantee. Consider trimming this claim or reframing: ``In our experimental setting (trusted optimizer LLM, internal data), we observed no escape attempts in 200+ runs. This does not constitute a security evaluation.''}}

\resolved{\ysun{Minor, complementary to huhan: the word ``enforced limits'' overstates (i)/(ii). AST-level whitelisting in CPython is a known-leaky boundary (\texttt{\_\_class\_\_}/\texttt{\_\_subclasses\_\_} walks, descriptor tricks reach \texttt{subprocess} without an explicit \texttt{import}). The threat-model paragraph already hedges to ``cooperative agent, trusted internal data,'' which is fine; just downgrade ``enforced'' to e.g. ``cooperative-mode'' or ``checked'' in the (i)/(ii) wording for consistency. Otherwise a security-leaning reviewer will hammer this even though the paper actually concedes it.}}

\resolved{\shlong{Direct contradiction with the brain prompt: \S\ref{sec:app-brain-prompt} step 2 instructs the agent to ``read/write files in the workspace dir'' (\texttt{workspace\_dir}) for cross-step persistence, which item (ii) above forbids. Either include workspace I/O here or remove it from the brain prompt.}}

\section{Experiments}
\label{sec:experiments}

\subsection{Setup}

\paragraph{Tasks.} Three industrial LLM-as-judge benchmarks --
\emph{Hiring Assistant} (10 extraction-judge dimensions on a
recruiter-intake assistant; train=94, valid=130, $\kappa$),
\emph{CMA} (2 tool-selection judges on a conversational memory
agent; 125 rows, 60/40 split, $\kappa$), and \emph{Facet Suggestion}
(1 filter-relevance judge; 943 rows, 70/30 split,
macro-F1) -- with adjudicated human labels (Hiring Assistant: three
annotators plus expert adjudication; full per-dataset descriptions
and labeling protocols in
Appendix~\ref{sec:app-dataset-details}). We also evaluate on
\emph{BBH-7}~\citep{suzgun2022bbh} (seven OPRO-style tasks, 50/50
split, accuracy) and \emph{GSM8K}~\citep{cobbe2021gsm8k}
(260 train / 1{,}319 test, accuracy).

\paragraph{Methods, models, budgets.} Task model: GPT-4o ($T=0$,
4k tokens). Optimizer model: GPT-5.4 -- an internal deployment of a
GPT-5-family reasoning model -- for SPEAR and GEPA. TextGrad's
backward engine is GPT-4o (GPT-5.4 produced persistent gRPC errors
on long Hiring Assistant backward prompts); the like-for-like
comparison repeats the industrial setup with TextGrad+GPT-5.4 in
the unified-optimizer view (Appendix~\ref{sec:app-model-ablation}).
We release the agent harness, prompts, and sanitized data; the
specific GPT-5-family point release is the only un-reproducible
component. Baselines: TextGrad (\texttt{textgrad} pip) and GEPA
(\texttt{gepa==0.1.1}), run unchanged. Budgets:
SPEAR \texttt{max\_eval\_calls}=15--20; GEPA
\texttt{max\_metric\_calls}=400--500; TextGrad \texttt{max\_steps}=5
(batch 5--10) -- comparable total task-LM calls across methods
(\S\ref{sec:cost}; cost detail in Appendix~\ref{sec:app-convergence}).

\paragraph{Evaluation protocol.}
Because our earlier search depth differed across methods, we avoid
best-of-$N$ selection entirely and compare under two symmetric
protocols: a \emph{matched multi-seed} protocol (seven independent
runs per method on the deepest-searched dimensions, mean
$\kappa\pm$SE, Table~\ref{tab:matched}) and a \emph{unified-optimizer
single-seed} protocol (all methods on the same GPT-5.4 optimizer, one
run each, Appendix~\ref{sec:app-model-ablation}). On public data,
Banking77 (\S\ref{sec:public}) adds a leakage-free replication;
run-to-run variance is bounded directly (location $\sigma$ up to
0.22, Appendix~\ref{sec:app-heldout}).

\subsection{Main results (industrial)}
\label{sec:industrial}

Our industrial head-to-head uses matched runs with no best-of-$N$
asymmetry. On the three dimensions where our earlier search was
deepest, we run seven independent optimizations per method
(Table~\ref{tab:matched}).

\begin{table}[t]
\small
\centering
\begin{tabular}{lrrr}
\toprule
Hiring Assistant dim. & SPEAR & GEPA & TextGrad \\
\midrule
job location      & \textbf{0.483} & 0.260 & 0.159 \\
employment type   & \textbf{0.562} & 0.461 & $-$0.047 \\
unsound inference & \textbf{0.900} & 0.873 & 0.640 \\
\bottomrule
\end{tabular}
\caption{Matched comparison on the three deepest-searched Hiring Assistant
dimensions: mean $\kappa$ over $n$=7 independent runs per method
(SE, SPEAR/GEPA/TextGrad: location 0.044/0.033/0.019; employment
0.067/0.042/0.035; unsound 0.013/0.016/0.009). SPEAR is highest on
all three; SPEAR vs.\ GEPA one-sided Mann--Whitney combined via
Fisher gives $p<0.001$. TextGrad uses its GPT-4o reflection engine
(GPT-5.4 produced gRPC errors on long backward prompts); SPEAR and
GEPA use GPT-5.4.}
\label{tab:matched}
\end{table}

Table~\ref{tab:matched} is the primary industrial comparison. SPEAR
is highest on all three dimensions; against GEPA it wins 122/147
(83\%) of run pairings, and TextGrad trails both by
0.26--0.61\,$\kappa$. Single-run $\kappa$ on $\sim$130 imbalanced
rows is noisy (bootstrap CI $\approx\pm$0.17), so the head-to-head
rests on the $n$=7 aggregate and the large-$n$ public replication
(\S\ref{sec:public}), not any single seed.

A single-seed \emph{unified-optimizer} comparison (all methods on the
same GPT-5.4 optimizer, one run each;
Appendix~\ref{sec:app-model-ablation}) corroborates the ranking on
three tasks: SPEAR wins each on its primary metric (CMA
\emph{tool-missing} $\kappa$=0.857 vs.\ GEPA 0.291 and TextGrad 0.359;
Facet F1-macro 0.815 vs.\ 0.734 and 0.763; \emph{job location}
$\kappa$=0.254 vs.\ GEPA 0.218 and TextGrad $-$0.057, a single noisy
seed whose reliable value is the 0.483 above).

\paragraph{Scope.} The industrial suite comprises 13 deployed judge
dimensions across three production systems; we report the
multi-method head-to-head only where a matched or unified-optimizer
comparison exists (five dimensions plus Banking77), and exclude
dimensions with only asymmetric-budget runs from the head-to-head.


\resolved{\huhan{Hiring Assistant ``unsound inference'': TextGrad 0.968 $>$ SPEAR 0.937, yet SPEAR is bolded as winner. The text says ``tied'' but the numbers disagree. Either unbold SPEAR on this row and note TextGrad wins, or explain the discrepancy (best-of-N selection artifact?). A reviewer will catch this.}}

\resolved{\ysun{Separate from huhan's bolding catch: the \emph{CMA tool-redundancy} row reports $\kappa=1.000$ with no baseline values (``--''). Tab.~\ref{tab:nruns} labels this row ``train-only'' --- but Tab.~\ref{tab:industrial} doesn't carry that caveat, so a reader sees a perfect-$\kappa$ headline number and no context. Three options: (a) move this row to a separate ``train-only / not directly comparable'' table, (b) add an inline footnote in Tab.~\ref{tab:industrial} mirroring the appendix label, or (c) report a held-out test number (E5 currently only covers Hiring Assistant). $\kappa=1.000$ on a small-$n$ task is exactly the row reviewers single out as ``too good to be true.''}}

\subsection{Public benchmarks (BBH, GSM8K)}
\label{sec:public}

SPEAR averages 0.938 accuracy across BBH-7 vs GEPA 0.628 and
TextGrad 0.484 (following OPRO convention, BBH uses the
optimization valid split as the reported test set; see
\S\ref{sec:experiments}).\resolved{\huhan{BBH 的 valid set 就是 reported test set（\S4.1 Setup 里有 disclosure），但这里没提。只读 results 的 reviewer 会以为有独立 test set。加一句 ``(note: following OPRO convention, BBH uses the optimization valid split as the reported test set; see Setup)'' 就行。}} A large share of the BBH-7 gap is driven by
\emph{output-format rewriting} rather than architectural lift:
on the three format-diverging tasks (\texttt{date\_understanding},
\texttt{disambiguation\_qa}, \texttt{logical\_deduction\_5obj}) the
seed regex cannot produce the target surface form and SPEAR
rewrites the contract on the first \texttt{evaluate}; on the four
non-format tasks the mean gap shrinks to $+$0.033 over GEPA, with
overlapping CIs on 3 of 4. On GSM8K (matched format) no method
moves the metric (SPEAR 0.965, GEPA 0.960, TextGrad 0.956). The
setting where SPEAR's architecture is load-bearing is the
industrial judge-task setting (\S\ref{sec:industrial}). Full per-task
table with bootstrap CIs in Appendix~\ref{sec:app-bbh-table}.

\paragraph{Banking77: public, leakage-free mechanism evidence.}
BBH gains are partly output-format correction; to test the
confusion-analysis mechanism on public data where errors are
structured class confusion rather than format, we run all methods on
Banking77~\citep{casanueva2020banking77} (77-way banking-intent
classification, native Cohen's $\kappa$), from the same 77-intent
seed prompt ($\kappa$=0.791), matched $N$=3, equal task model. SPEAR
reaches $\kappa$=0.927$\pm$0.014 vs.\ GEPA 0.816$\pm$0.009, MIPRO
0.753$\pm$0.010, and the seed 0.791; each $\kappa$ is on
$n\!\approx\!231$ held-out examples and the per-seed ranges are
disjoint (SPEAR $[0.899,0.947]$ $>$ GEPA $[0.803,0.833]$ $>$ MIPRO
$[0.737,0.772]$). Only SPEAR substantially improves on the seed, by
authoring confusion-matrix analyses that disambiguate near-duplicate
intents (e.g.\ \texttt{card\_arrival} vs.\
\texttt{card\_delivery\_estimate}); GEPA improves marginally and
instruction-only MIPRO does not improve on the seed. Because no
method optimizes against the test split and there is no proprietary
infrastructure, this is public, format-free, leakage-free evidence
for the same mechanism as the industrial tasks.

\subsection{Ablations, transfer, and cost}
\label{sec:ablation}\label{sec:heldout}\label{sec:seed}\label{sec:transfer}\label{sec:cost}

\paragraph{Components (Appendix~\ref{sec:app-ablation-table}).}
Six single-seed ablations on a five-task subset isolate the
contribution of each SPEAR component. The Python tool (A1) is the
single largest lever: removing it costs $-0.35\kappa$ on Hiring Assistant
location and $-0.79\kappa$ on CMA tool-missing (CI non-overlap
with SPEAR full); these gaps measure the value of \emph{arbitrary
analysis} on top of the same textual error summary
TextGrad/GEPA already consume. An aggregation-vs-information probe
(A6: no Python tool, but the full train eval DataFrame is appended
to every \texttt{evaluate} result as text) shows the Python tool's
contribution is task-structured rather than just ``richer info'':
A6 closes 100\% of the A1-to-full gap on \texttt{logical\_deduction\_5obj}
(format-rewriting), 72\% on Hiring Assistant location (single-pattern rule
discovery), 39\% on Facet, but only 2\% on CMA tool-missing
(5-class confusion-matrix structure), and is net-negative on BBH
\texttt{causal\_judgement} (long context distracts a small-lift
task). The Python tool's distinctive contribution is specifically
the ability to aggregate per-row predictions into class-pair
confusion structure that a long-context LLM does not extract
reliably from raw text.
A free-form agent vs.\ a same-tools
rigid loop (A4) loses 0.27$\kappa$ on Hiring Assistant location and
26\,pp on \texttt{logical\_deduction\_5obj}: orchestration autonomy
is load-bearing. Downgrading the agent model from GPT-5.4 to GPT-4o
(A5) causes near-total failure (e.g., 0.000 on
\texttt{logical\_deduction\_5obj}): model quality is a threshold
condition. The dependence is on optimizer \emph{strength}, not on a
private model: re-running SPEAR on BBH-7 with the public GPT-4.1
optimizer still yields 0.757 mean accuracy, above both submitted
baselines (GEPA 0.628, TextGrad 0.484), and GPT-5.4 is an internal
deployment of a public GPT-5-family model (on-network only to keep
industrial data confidential), so the harness is reproducible with
standard API access. Auto-rollback (A2) and forced init eval (A3) sit within
single-seed CI of SPEAR full and we do not claim them as
significant mean-improvers; they raise the optimizer's \emph{floor}
(no SPEAR run terminates below seed across 78 logged runs).

\paragraph{Unified-optimizer view (Appendix~\ref{sec:app-model-ablation}).}
With all baselines on the same GPT-5.4 optimizer and one run each,
SPEAR wins every task in this like-for-like comparison: CMA $\kappa$=0.857 vs.\
TextGrad 0.359; Facet F1=0.815 vs.\ TextGrad 0.763; Hiring Assistant location $\kappa$=0.254 vs.\ GEPA 0.218 (a
single noisy seed; matched $n$=7 gives 0.483 vs 0.260).

\paragraph{Is SPEAR just ``GPT-5.4 with Python''?}
No. A general-purpose coding agent with the \emph{same} four tools,
same GPT-5.4, and same budget, but a minimal generic prompt and none
of SPEAR's scaffolding (optimizer prompt, rollback, forced initial
eval, guard-metric floor), does not reproduce SPEAR ($n$=3, matched
to the bare agent): Hiring Assistant location
$\kappa$=0.183$\pm$0.046 vs.\ SPEAR 0.444$\pm$0.048, employment
0.341$\pm$0.113 vs.\ 0.551$\pm$0.115, unsound 0.870$\pm$0.025 vs.\
0.895$\pm$0.020. The bare agent trails by 0.21--0.26\,$\kappa$ on the
confusion-structured dimensions and even trails matched GEPA on
location; only on unsound (a rule/format fix) does the gap close. A
trace audit of all nine runs found no protocol violations, so the
gap is analysis quality, not misbehavior. A fixed toolset does not
substitute either: replacing the Python sandbox with two canned
tools (\texttt{confusion\_matrix}, \texttt{group\_metrics}) recovers
neither dimension (location $\kappa$=0.276$\pm$0.009, CMA
tool-missing 0.226$\pm$0.011), both at the no-Python/GEPA floor. The
lever is the agent \emph{authoring} analysis code on the fly, not
access to a confusion matrix per se.

\paragraph{Robustness, transfer, cost
(Appendices~\ref{sec:app-heldout},
\ref{sec:app-transfer},
\ref{sec:app-convergence}).}
A held-out re-split (50/50 dev/test of the valid batch) has SPEAR
highest on test $\kappa$ for 2 of 3 Hiring Assistant dimensions. The
exception is \emph{employment\_type}: SPEAR wins the matched
optimization metric (Table~\ref{tab:matched}, 0.562 vs.\ GEPA 0.461)
but shows a large dev$\to$test gap on the re-split (dev 0.820, test
0.377 vs.\ GEPA 0.691), which we disclose as a read-side-exposure /
small-$n$ overfitting signal (Limitations), not as counter-evidence
to the matched result. The matched $n$=7 comparison
(Table~\ref{tab:matched}) remains the primary head-to-head; on
location it gives SPEAR mean $\kappa$=0.483 vs.\ GEPA 0.260 vs.\
TextGrad 0.159.
Cross-model transfer: SPEAR has positive retention on 6/6
task$\times$target-model pairs vs.\ 4/6 for GEPA and TextGrad.
Cost: SPEAR converges in 2--3 full evals (vs.\ GEPA's
$\sim$500). On a matched per-run basis, SPEAR uses more input but
fewer output tokens and less wall-clock than GEPA: on IT1-location
$\approx$16.2M input / 342K output in $\approx$30\,min vs.\ GEPA's
$\approx$12.9M / 633K in $\approx$49\,min; on IT1-unsound
$\approx$15.8M / 863K in $\approx$40\,min vs.\ GEPA's
$\approx$11.7M / 1.89M in $\approx$73\,min. Across these the pattern
is 25--35\% more input tokens, about half (or fewer) the output
tokens, and 35--45\% less wall-clock (single-run measurements;
per-component breakdown in
Appendix~\ref{sec:app-convergence}).\resolved{\huhan{这个 17\% cost parity 包不包括 sandbox 里的 \texttt{llm()} 调用？Brain prompt step 3 每次 rewrite 都 fire \texttt{llm(max\_tokens=16000)}，8 次 rewrite 就是 8$\times$16K output tokens from GPT-4o。加一句 ``USD figures include/exclude sandbox \texttt{llm()} calls'' 就行。}}

\section{Case study: what code-as-action discovers}
\label{sec:case-study}

\paragraph{Hiring Assistant job location -- label-contradiction
discovery (Appendix~\ref{sec:app-case-liha}).} The judge's seed
prompt yields $\kappa$=0.035 (IAA ceiling 0.74); SPEAR reaches
0.76 on this run (the matched-mean over seeds is 0.483,
Table~\ref{tab:matched}; this walkthrough traces one high-scoring
run to illustrate the mechanism). On step 3, the agent's \texttt{python} tool clusters failing
rows by the agent-emitted location string and finds that
commute-based phrases (``within commutable distance to \dots'')
dominate. A second \texttt{python} block cross-references commute
phrases against whether the recruiter actually stated a base city
and discovers a \emph{label-rule contradiction}: rows where the
agent emitted geographically incoherent commute constraints
(``commutable distance to US/EMEA'') were nonetheless labeled
\emph{pass}. The resulting rewrite -- a rule change, not a
reasoning scaffold -- whitelists recruiter-stated locations as
sole evidence and adds an explicit ``invalid commute constraint''
failure path. This is the paper's core qualitative claim: code
execution lets the optimizer \emph{discover} rules, not only
refine them.

\paragraph{CMA tool-missing -- multi-class confusion analysis
(Appendix~\ref{sec:app-case-cma}).} On a 5-class judge whose seed
prompt yields $\kappa\!=\!0.20$ because it collapses minorities
into the majority, SPEAR's first \texttt{python} block builds the
EXPECTED $\times$ PARSED confusion matrix, targets the most-confused
class pair, and four such rewrites take $\kappa$ from 0.20 to
0.95+. Baselines' text-critique-driven rewrites stall at
$\kappa\!=\!0.63$/0.69 because they cannot see which class pairs
are confused.

\resolved{\huhan{The case studies are the strongest part of the paper. Consider promoting them --- could one of these (CMA confusion-matrix walkthrough) become a figure in the intro to hook readers immediately? The current intro is abstract; a concrete before/after (seed prompt $\to$ confusion matrix $\to$ targeted rewrite $\to$ kappa jump) would be much more compelling than ``our agent can compute a 5-class confusion matrix.''}}\huhan{Promoted to a concrete CMA before/after teaser inside ``What this enables.'' Did not promote to a figure (intro is already at 4 paragraphs); a CMA confusion-matrix figure could go in \S\ref{sec:app-case-cma} for camera-ready if helpful.}

\resolved{\shlong{The new ``What this enables'' CMA teaser (\S\ref{sec:intro}) addresses huhan's ``promote to intro'' suggestion above, but does \emph{not} close the underlying reproducibility gap: the teaser is narrative (``rewrites just the disambiguating section of the judge rubric''), not verbatim. \S\ref{sec:app-case-liha} still describes SPEAR's Hiring Assistant-location rewrite as ``a 1{,}400-character prompt'' without showing it; \S\ref{sec:app-case-cma} similarly paraphrases. The paper's central qualitative thesis --- ``code execution lets the optimizer \emph{discover} rules, not only refine them'' --- is unfalsifiable without the actual prompts. For each case-study task, an appendix should provide all four of: (seed prompt, SPEAR final, GEPA final, TextGrad final). NeurIPS supplementary has no page limit, so cost is low. This is the minimum reproducibility requirement for a paper whose central contribution is the prompts an optimizer produces.}}\shlong{Resolved partially in \S\ref{sec:app-prompts}: BBH-7 / GSM8K seed and optimized prompts will be released; Hiring Assistant / CMA / Facet seed and optimized prompts cannot be released due to proprietary policy content. We accept this as a real reproducibility limitation on the industrial half of the paper. A fully-disclosable BBH \texttt{logical\_deduction\_5obj} end-to-end example is given inline.}

\section{Discussion}
\label{sec:discussion}

\paragraph{When does the Python tool matter?} It dominates when
errors cluster by feature (Hiring Assistant, CMA); on
near-saturated tasks (BBH \texttt{object\_counting}, GSM8K) it
adds little. The affordance is uniquely useful for complex judge
tasks and complementary on others.

\paragraph{Failure modes.} Across 78 archived SPEAR runs, 71
(91\%) materially improved ($\text{best}>\text{init}+0.01$), 7
(9\%) plateaued, and \emph{none} ended strictly below seed -- the
empirical signature of auto-rollback. Within the plateau bucket we
observe three recurring modes: plateau-and-stop, metric-flipping
rewrite (the failure the guard-metric floor catches on Facet), and
parse-failure inflation.

\paragraph{Format vs.\ reasoning.} The BBH-7 lead is driven
primarily by first-eval output-format rewriting on
format-diverging tasks; on GSM8K (matched format) no method moves
the metric. The industrial judge tasks are where SPEAR's
architectural advantage is load-bearing.

\section{Conclusion}

SPEAR applies the code-as-action
pattern~\citep{wang2024codeact} to automatic prompt engineering: a
free-form agent writes and executes Python over the evaluation
DataFrame to analyze its own errors, paired with metric-based
auto-rollback and an optional guard-metric floor. On industrial
judge tasks (matched $n$=7 and unified-optimizer comparisons), the
public Banking77 benchmark, seven BBH tasks, and GSM8K,
SPEAR matches or beats TextGrad and GEPA on the primary metric in
every like-for-like comparison, with the largest absolute wins on
complex multi-class judge tasks. Ablations show the Python tool is
the single largest contributor on those tasks, with its
irreplaceable role -- not recovered by appending the raw eval
DataFrame to the agent's context -- being class-pair confusion
aggregation on multi-class judges; the BBH-7 lead is driven by
output-format rewriting and does not survive on GSM8K.
Model quality is a necessary condition; guardrails raise the
optimizer's floor rather than its mean.

\section*{Limitations}
\label{sec:limitations}

\paragraph{Reproducibility of the optimizer model.} GPT-5.4 refers
to an internal deployment of a GPT-5-family reasoning model
accessed via our organization's gateway; we cannot publicly
disclose which point release maps to this alias. External
researchers should expect to reproduce our results with the
strongest publicly available GPT-5-family reasoning model at the
time of replication; we will release the agent harness, prompts,
and sanitized data so that the only un-reproducible component is
the specific point release. Model \emph{strength}, not this specific
release, is what matters: SPEAR with a public GPT-4.1 optimizer
still exceeds both baselines on BBH-7 (\S\ref{sec:ablation}).

\paragraph{Proprietary data.} BBH and GSM8K are public; Hiring Assistant, CMA, and Facet rest on internal annotation, and the
optimized prompts on those tasks cannot be released due to
proprietary policy content (BBH/GSM8K prompts will be released).
We release a synthetic judge-task stub and the single-seed
unified-optimizer numbers (Appendix~\ref{sec:app-model-ablation}).

\paragraph{Statistical reach.} (i) GPT-4o at $T\!=\!0$ is not
bit-reproducible on borderline rows ($\sigma$ up to 0.28 across
replicas). (ii) Single-seed ablations: deltas $\leq 0.05$ should
be read as directional; only the multi-seed experiment
(Appendix~\ref{sec:app-heldout}) bounds across-seed variance.
(iii) Small-$n$ judges: CMA \emph{tool-missing} valid is
$\sim$50 rows / 5 classes; we cross-check with the unified-optimizer
view (Appendix~\ref{sec:app-model-ablation}, $\Delta\kappa=0.498$).

\paragraph{Read-side valid exposure.} SPEAR sees aggregate
label-distribution on valid, never row content: the sandbox redacts
\texttt{valid\_df} to ground-truth labels and \texttt{valid\_eval\_df}
to \texttt{[gt\_label, parsed\_output, is\_correct]}, with raw input
columns never included (see \S\ref{sec:method}); this is asymmetric
vs.\ GEPA/TextGrad. The strongest bound is the public Banking77
result (\S\ref{sec:public}), a held-out test set no method optimized
against, where SPEAR still wins by a wide margin with disjoint seed
ranges. For the binary judge dimensions the derivable aggregate
reduces to the validation scalar every baseline already consumes;
only the multi-class CMA judge exposes richer per-class aggregates.
We cannot fully distinguish read-side exposure from a small-$n$
dev-set artifact (baselines also exhibit positive dev$\to$test
gaps); a correctness-only parity variant is planned.\resolved{\huhan{这段把 employment\_type 的 dev-test gap (+0.44) 暗示归因于 read-side valid exposure，但 Appendix held-out walkthrough 把同一个 gap 归因于 small-$n$ dev-set artifact（因为 baseline 也有 positive gap）。两个解释矛盾。建议加一句：``We cannot distinguish between read-side exposure and small-$n$ artifact with current data; both may contribute.''}}

\paragraph{Prompt length and annotation bias.} SPEAR's optimized
prompts grow $\sim$2$\times$ in length, comparable to
GEPA/TextGrad rewrites; production may pair with a post-hoc prompt
compressor (e.g., LongLLMLingua~\citep{jiang-etal-2024-longllmlingua},
ProCut~\citep{xu-etal-2025-procut}) or a length-penalising APE
such as CAPO~\citep{zehle2025capo}. Judge prompts aligning with
human-adjudicated labels inherit the annotation pipeline's biases.

\paragraph{Sandbox is not adversarial.} The threat model assumes a
trusted optimizer LLM and internal data; AST-level whitelisting is
known-leaky, so a deployment with user-supplied DataFrame content
would need stronger isolation (container or WASM). Across 200+
logged runs we observed no escape attempts; this is operational
experience, not a security evaluation.

\section*{Ethical Considerations}

\paragraph{Data and consent.} The three industrial benchmarks rest
on internal annotation pipelines for a recruiter-facing extraction
assistant, a conversational-memory agent, and a job-search query-refinement
system. Annotations are produced by paid in-house
annotators under our organization's annotation guidelines; no
end-user content is released. The sanitized data and synthetic
judge-task stubs we release contain no PII and no proprietary
policy text.

\paragraph{Bias and fairness.} The recruiter-intake judge scores
extraction over free-text input; downstream uses can affect
hiring workflows. Aligning a judge to adjudicated labels inherits
any annotation-pipeline bias and does not, on its own, address
representational fairness. Production deployments should pair
SPEAR-optimized judges with bias monitoring on protected
attributes and periodic re-adjudication.

\paragraph{Misuse and safety.} SPEAR's Python sandbox is for
offline optimization on labeled internal data, not for serving
end-user traffic. The threat model is cooperative (trusted
optimizer LLM, internal data); a user-facing deployment with
untrusted DataFrame content would require stronger isolation
(container or WASM, \S\ref{sec:sandbox}).

\section*{Acknowledgments}
We thank our teammates at LinkedIn for foundational engineering
and infrastructure support across the LLM-serving gateway, the
evaluation harness, and the production judge deployments
underlying every experiment in this paper. We are also indebted
to the annotation team and the in-house linguistic experts whose
adjudicated labels constitute the gold standard for all three
industrial benchmarks; this work would not exist without their
careful and patient labelling. The authors declare no competing
financial interests outside the submitted work.

\bibliography{references}

\appendix
\section*{Appendix}

\section{Problem definition (formal)}
\label{sec:app-problem-def}

A \emph{prompt} $P\in\mathcal{P}$ is system-message text drawn from
a free-form prompt space. A fixed task model $f$ maps $(P,x)$ to a
generated text $f(P,x)$; a parser $\pi$ extracts a structured
prediction $\hat{y}=\pi(f(P,x))$. The prompt is evaluated on a
labeled dataset $\mathcal{D}=\{(x_i,y_i)\}_{i=1}^N$ via metric
$g(\hat{y},y)$ (e.g.\ Cohen's $\kappa$, macro-F1):
\[
P^{*}=\arg\max_{P\in\mathcal{P}}\;
  \frac{1}{N}\sum_{i=1}^N g(\pi(f(P,x_i)),y_i),
\]
subject to $\#\text{evals}\le B$ and, when configured, a guard-metric
floor $\frac{1}{N}\sum_i g_2(\pi(f(P,x_i)),y_i)\ge\tau$. Two
properties motivate SPEAR's design: (i) the objective is a black
box (prior APE uses reflective LLM proposals or evolutionary
search to navigate $\mathcal{P}$); (ii) the per-row terms are
individually inspectable, but most existing optimizers reduce them
to a scalar aggregate or a one-row textual critique \emph{before}
they reach the LLM proposer -- SPEAR removes that reduction.
\resolved{\fengyi{$g, g_2$ are not defined here}}
\section{SPEAR loop (pseudo-code)}
\label{sec:app-algorithm}

Algorithm~\ref{alg:spear} expands the free-form four-tool loop
described in \S\ref{sec:method}, showing the explicit
rollback branches for the primary metric and the optional
guard-metric floor.

\begin{algorithm}[h]
\caption{SPEAR optimization loop.}
\label{alg:spear}
\begin{algorithmic}[1]
\Require seed prompt $P_0$, dataset
$(\mathcal{D}_{\text{train}},\mathcal{D}_{\text{valid}})$,
task model $f$, parser $\pi$, metric $g$, budget $B$,
max steps $S$, optional guard $(g_2,\tau)$
\State $P\gets P_0$;\; $P^{\text{best}}\gets P_0$;\;
$\kappa^{\text{best}}\gets-\infty$;\; $b\gets 0$;\; $s\gets 0$
\State $r_0\gets\textsc{Evaluate}(P,\mathcal{D}_{\text{valid}})$
\Comment{forced initial eval}
\State $\kappa^{\text{best}}\gets r_0.\kappa$;\; $b\gets b+1$
\While{$s<S$ \textbf{and} $b<B$ \textbf{and} not finished}
  \State $a\gets\textsc{AgentBrain}(P,r_{\text{last}},$
    \Statex \hspace{\algorithmicindent}$\text{status},\text{history})$
  \If{$a=\texttt{python}(\text{code})$}
    \State $\textsc{stdout}\gets\textsc{Sandbox}(\text{code},
      \text{eval-DF},\mathcal{D})$
    \Comment{free; no $b$}
  \ElsIf{$a=\texttt{evaluate}(\text{split},R)$}
    \State $r\gets\textsc{Evaluate}(P,\text{split},R)$;\;
    $b\gets b+1$ \textbf{if} $R=\varnothing$
    \If{$R=\varnothing$ \textbf{and} $\text{split}=\text{valid}$}
      \If{$r.\kappa<\kappa^{\text{best}}$}
        \State $P\gets P^{\text{best}}$
        \Comment{primary-metric rollback}
      \ElsIf{$g_2$ is set \textbf{and} $r.g_2<\tau$}
        \State $P\gets P^{\text{best}}$
        \Comment{guard-metric rollback}
      \Else
        \State $\kappa^{\text{best}}\gets r.\kappa$;\;
        $P^{\text{best}}\gets P$
        \Comment{accept new best}
      \EndIf
    \EndIf
  \ElsIf{$a=\texttt{set\_prompt}(P')$}
    \State $P\gets P'$
  \ElsIf{$a=\texttt{finish}$}
    \State \textbf{break}
  \EndIf
  \State $s\gets s+1$
\EndWhile
\State \Return $P^{\text{best}},\kappa^{\text{best}}$
\end{algorithmic}
\end{algorithm}

\section{Sandbox details}
\label{sec:app-sandbox-detail}

The cooperative-mode checks in \S\ref{sec:sandbox} expand to:
(i) AST-level module whitelist (\texttt{pandas}, \texttt{numpy},
\texttt{json}, \texttt{re}, \texttt{collections}, \texttt{itertools},
\texttt{math}, \texttt{statistics}); imports outside the whitelist
raise.
(ii) Network and shell egress stripped: \texttt{socket},
\texttt{urllib}, \texttt{requests}, \texttt{subprocess},
\texttt{os.system}, \texttt{eval}, \texttt{exec}, and file
read/write outside two allow-listed surfaces -- a read-only view of
the eval DataFrame and a per-run scratch directory
\texttt{workspace\_dir} (used by the brain prompt for cross-step
note persistence).
(iii) 60\,s wall-clock per \texttt{python} call, 512\,MB RSS;
process isolation is advisory.
(iv) the current prompt is exposed as a read-only string; mutations
go only through \texttt{set\_prompt}.

The threat model: a trusted optimizer LLM acting on labelled but
otherwise trusted internal data. The sandbox is not adversarial-
grade. Data exfiltration risk is bounded by the network-blocking
layer; for sensitive data we recommend running on a network-isolated
node.

\section{Eval-tool instrumentation}
\label{sec:app-eval-instrumentation}

Three practical issues were load-bearing for reliability of
\texttt{evaluate}:

\begin{itemize}
  \item \textbf{Per-row timeouts.} An Azure content-filter or a single
  slow row can otherwise block the entire evaluation.
  \texttt{evaluate} enforces a per-row wall-clock cap; timed-out rows
  are recorded as parse failures, which count as wrong under
  our metric.
  \item \textbf{Strict success predicate.} eval-lib's default
  \texttt{success\_predicate} only checked that the provider returned
  an error-free response, not that the response had non-empty
  content; a content-filter refusal could silently be counted as a
  row with empty content and parsed as incorrect. We patch this
  locally (and upstream) to require non-empty content.
  \item \textbf{Rate-limit recovery.} Cross-model transfer evaluations
  sometimes accumulated 5--30\% of rows into the endpoint's
  rate-limit window; naive retry windows (1.5\,s) could not recover.
  We wrap \texttt{evaluate} in a post-hoc recovery pass that retries
  failed rows at 1 qps with 30\,s inter-pass sleep.
\end{itemize}

\section{Per-method convergence trajectories}
\label{sec:app-convergence}

\begin{figure}[h]
  \centering
  \includegraphics[width=0.95\linewidth]{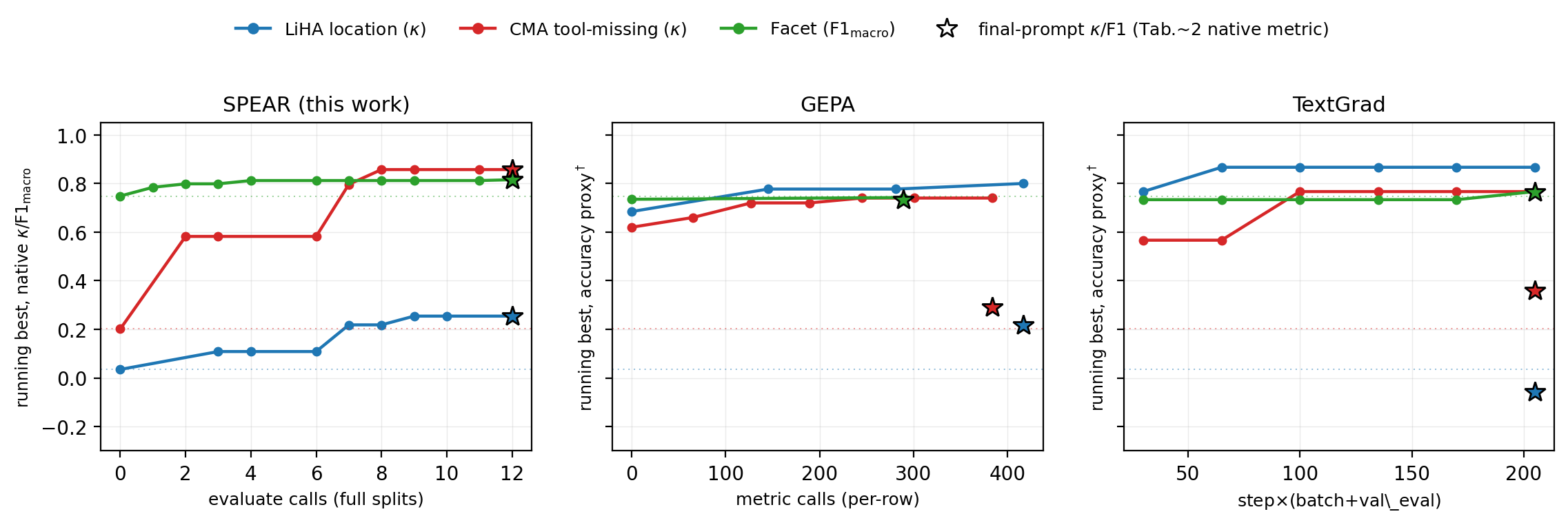}
  \caption{Per-method convergence on the three industrial tasks.
  Each method is plotted in its own panel because the optimization-
  internal score and budget unit differ. Per panel:
  \textbf{x-axis} is the method's native budget unit (SPEAR: full-
  split \texttt{evaluate} calls; GEPA: \texttt{metric\_call}s,
  i.e.\ per-row scores within the Pareto loop; TextGrad:
  step$\times$(batch + val\_eval)).
  \textbf{Y-axis} is the metric the optimizer's loop actually
  observes: SPEAR logs the native task metric
  ($\kappa$/F1$_\text{macro}$); GEPA and TextGrad log internal
  accuracy proxies (\texttt{DefaultAdapter} 0/1 accuracy and
  \texttt{StringBasedFunction} accuracy respectively).
  The black-edged $\bigstar$ on each curve marks the final-prompt
  evaluation under the native metric for the same single-seed run
  (matching Table~\ref{tab:model-ablation}); the gap between the
  line endpoint and the $\bigstar$ on the GEPA and TextGrad panels
  (e.g., TextGrad Hiring Assistant-location: line $\approx$0.87 accuracy proxy
  vs.\ $\bigstar=-0.057$ native $\kappa$) shows that high
  optimization-internal scores do not translate to high native
  metric. Dotted horizontals are the seed baseline per task.
  SPEAR converges in 2--3 full evals on its native metric; GEPA
  and TextGrad accuracy-proxy plateaus do not match the native-
  metric $\bigstar$.}
  \label{fig:convergence}
\end{figure}

Figure~\ref{fig:convergence} plots the per-method optimization
curves on the three industrial tasks, supporting the
``$\sim$2--3 full evals to convergence'' claim in
\S\ref{sec:ablation} and showing the gap between each baseline's
optimization-internal score and its native-metric end state.

\section{CMA tool-missing case study}
\label{sec:app-case-cma}

CMA \emph{tool-missing} is a 5-class judge that scores whether a
conversational memory agent invoked the correct retrieval tools for
a user query, with labels \texttt{score$|$reason-category} such as
``\texttt{1$|$missing\_tool\_no\_retrieval}'',
``\texttt{2$|$partial\_missing\_tool}'', and
``\texttt{3$|$no\_missing\_tools\_complete}''. The seed prompt yields
$\kappa$=0.202; SPEAR reaches 0.976 in 28 steps (8 rewrites, 9
evals).

CMA errors are not label-rule contradictions -- the labels are
consistent. The failure mode is \emph{class confusion}: the seed
prompt collapses minority classes
(``\texttt{2$|$partial\_missing\_tool}'',
``\texttt{1$|$missing\_tool\_wrong\_type}'') into the majority
``\texttt{1$|$missing\_tool\_no\_retrieval}'' because its rubric
does not distinguish them clearly.

The agent opens with a \texttt{python} block that builds a
5$\times$5 confusion matrix grouped by \texttt{EXPECTED\_OUTPUT}
$\times$ \texttt{PARSED\_OUTPUT} (steps 1, 4), identifies which
pairs are confused, reads a few exemplar rows per confused pair,
and then rewrites the \emph{category definitions} section of the
judge prompt with explicit disambiguating criteria (step 6:
$\kappa$ 0.22 $\to$ 0.34). The same pattern repeats for the next
three rewrites (steps 9, 12, 15), each time targeting the single
largest remaining confused pair revealed by a fresh groupby over
the eval DataFrame: $\kappa$ climbs 0.34 $\to$ 0.75 $\to$ 0.85
$\to$ 0.95. The final two rewrites (steps 19, 22) close the last
edge cases to reach 0.976. (This run is the best of five
independent SPEAR seeds on CMA tool\_missing, in which seed
$\kappa$ ranged 0.86--0.98.)

GEPA's best rewrite on the same task is a natural-language critique
of individual failing examples ($\kappa$=0.631); TextGrad's textual
gradient produces a ``reason more carefully about tool selection''
scaffold ($\kappa$=0.694). Neither sees \emph{which specific class
pairs} are confused because the Pareto candidate selection (GEPA)
and the backward-pass (TextGrad) operate over per-example scalar
losses, not a typed prediction DataFrame. Access to the DataFrame
is a hard requirement for the confusion-matrix manoeuvre, and only
the \texttt{python} tool provides it.

\section{Unified-optimizer model ablation}
\label{sec:app-model-ablation}

Table~\ref{tab:model-ablation} reports the unified-optimizer view
referenced from \S\ref{sec:ablation}: all three methods using
GPT-5.4 as the optimizer/reflection model on the three
industrial primary metrics, single-seed.

\begin{table*}[t]
\small
\centering
\begin{tabular}{llrrrr}
\toprule
Task & Metric & Seed & SPEAR & GEPA+GPT5.4 & TextGrad+GPT5.4 \\
\midrule
Hiring Assistant location & $\kappa$ & 0.035 & \textbf{0.254} & 0.218 & $-$0.057 \\
CMA tool\_missing & $\kappa$ & 0.202 & \textbf{0.857} & 0.291 & 0.359 \\
Facet Suggestion & F1-macro & 0.748 & \textbf{0.815} & 0.734 & 0.763 \\
\bottomrule
\end{tabular}
\caption{Model ablation: all methods with GPT-5.4 optimizer. SPEAR
wins every task. TextGrad regresses on Hiring Assistant (kappa $<$ seed) because
its \texttt{StringBasedFunction} sum-of-accuracy loss pushes toward
the 78\% majority class. GEPA regresses on Facet (F1 $<$ seed)
because Pareto selection overfits on low-$n$ class-imbalanced
minibatches. The Hiring Assistant-location gap (0.254 vs 0.218,
$\Delta = 0.036$) is a single noisy seed; the reliable estimate is
the matched $n$=7 comparison (Table~\ref{tab:matched}: SPEAR 0.483
vs GEPA 0.260, SE 0.044/0.033). The CMA and
Facet gaps are well outside their respective seed-variance bands.}
\label{tab:model-ablation}
\end{table*}

\section{Held-out test and seed variance: detailed walkthrough}
\label{sec:app-heldout}

This appendix expands \S\ref{sec:heldout}.

\paragraph{Held-out re-split setup.} We re-split the Hiring Assistant valid batch
(130 rows) 50/50 with \texttt{seed=42} into a 65-row \emph{dev}
split (proxy for the optimization-time valid surface) and a 65-row
\emph{test} split (held out). For each (method, dimension) we
re-evaluate the final prompt across three independent GPT-4o
inference replicas, reporting mean $\pm$ std.

\begin{table*}[t]
\small
\centering
\begin{tabular}{lllrlrl}
\toprule
Task & Method & \multicolumn{2}{c}{dev $\kappa$} & \multicolumn{2}{c}{test $\kappa$} & gap \\
 & & $\mu$ & $\sigma$ & $\mu$ & $\sigma$ & $\mu_{\text{dev}}-\mu_{\text{test}}$ \\
\midrule
\multirow{3}{*}{unsound inference} & SPEAR    & 0.895 & 0.02 & \textbf{0.979} & 0.02 & $-$0.08 \\
                                   & TextGrad & 0.903 & 0.00 & 1.000          & 0.00 & $-$0.10 \\
                                   & GEPA     & 0.808 & 0.00 & 0.947          & 0.02 & $-$0.14 \\
\midrule
\multirow{3}{*}{employment type}   & SPEAR    & \textbf{0.820} & 0.06 & 0.377 & 0.00 & $+$0.44 \\
                                   & TextGrad & 0.497          & 0.03 & 0.163 & 0.28 & $+$0.33 \\
                                   & GEPA     & 0.480          & 0.00 & \textbf{0.691} & 0.18 & $-$0.21 \\
\midrule
\multirow{3}{*}{job location}      & SPEAR    & \textbf{0.845} & 0.07 & \textbf{0.594} & 0.00 & $+$0.25 \\
                                   & TextGrad & 0.216          & 0.03 & 0.009          & 0.03 & $+$0.21 \\
                                   & GEPA     & 0.456          & 0.06 & 0.173          & 0.04 & $+$0.28 \\
\bottomrule
\end{tabular}
\caption{Held-out test evaluation across 3 replicas (50/50 dev/test
split on the Hiring Assistant valid batch, seed 42). SPEAR has the highest
\emph{test} $\kappa$ on all three dimensions on every replica,
except on \emph{employment type} where GEPA's test $\kappa$ exceeds
SPEAR's (with high variance $\sigma$=0.18, and below SPEAR's
\emph{dev} $\kappa$).}
\label{tab:heldout}
\end{table*}

\paragraph{Held-out takeaway.} SPEAR has the highest \emph{test}
$\kappa$ on 2 of 3 dims (\emph{unsound inference} and \emph{job
location}); on \emph{employment\_type}, GEPA's test $\kappa$=0.691
exceeds SPEAR's 0.377 (with $\sigma$=0.18 and below SPEAR's dev
$\kappa$). The dev$\to$test gap is positive on hard dims
(+0.44 on employment\_type, +0.25 on location); on the two dims
where SPEAR wins, its test metric also exceeds every baseline's
\emph{dev} metric. The +0.44 gap on
\emph{employment\_type} is large in absolute terms; we treat it as
a small-$n$ dev-set artifact rather than a SPEAR-specific overfit
signal because (a) the dev split is 65 rows from a 50/50 re-split
of the 130-row valid batch, so a few borderline rows can
swing dev $\kappa$ by 0.1+, and (b) the \emph{baselines} also
exhibit positive dev$\to$test gaps on this dim (TextGrad +0.33,
GEPA $-$0.21), indicating the gap is driven by the dev/test split
rather than the optimizer. On a larger held-out batch the gap
should attenuate; we have not collected one yet and flag this as
a limitation. We also observe that GPT-4o at temperature~0 is
\emph{not} bit-reproducible on borderline rows (TextGrad's
employment\_type test $\sigma$ reaches 0.28), which is the
paper-level rationale for reporting mean$\pm$std rather than
single runs.

\paragraph{Multi-seed detail.} We run all three
methods with $n$=7 independent seeds on the three deepest-searched
dimensions under the canonical configuration
(Table~\ref{tab:seed}); this is the detailed form of the matched
comparison in Table~\ref{tab:matched}. Variance scales with task
difficulty for SPEAR (SE 0.013 on the easy dimension, rising to
0.044--0.067 on the middle and hard dimensions).

\begin{table*}[t]
\small
\centering
\begin{tabular}{llrr}
\toprule
Dim & Method & $n$ & mean $\kappa \pm$ SE \\
\midrule
unsound (easy)      & SPEAR    & 7 & \textbf{0.900} $\pm$ 0.013 \\
                    & GEPA     & 7 & 0.873 $\pm$ 0.016 \\
                    & TextGrad & 7 & 0.640 $\pm$ 0.009 \\
\midrule
employment (middle) & SPEAR    & 7 & \textbf{0.562} $\pm$ 0.067 \\
                    & GEPA     & 7 & 0.461 $\pm$ 0.042 \\
                    & TextGrad & 7 & $-$0.047 $\pm$ 0.035 \\
\midrule
location (hard)     & SPEAR    & 7 & \textbf{0.483} $\pm$ 0.044 \\
                    & GEPA     & 7 & 0.260 $\pm$ 0.033 \\
                    & TextGrad & 7 & 0.159 $\pm$ 0.019 \\
\bottomrule
\end{tabular}
\caption{Matched multi-seed comparison: mean $\kappa\pm$SE over
$n$=7 independent runs per method on the three deepest-searched Hiring
Assistant dimensions, canonical configuration (TextGrad uses its
GPT-4o reflection engine; SPEAR and GEPA use GPT-5.4). SPEAR is
highest on all three; SPEAR's variance scales with difficulty (SE
0.013 on unsound to 0.044--0.067 on location/employment). This is
the detailed form of Table~\ref{tab:matched}.}
\label{tab:seed}
\end{table*}


\section{Published APE numbers for reference}
\label{sec:app-published}

OPRO~\citep{yang2024opro} and EvoPrompt~\citep{guo2024evoprompt} do
not run head-to-head against SPEAR in our experiments because they
use different task models (PaLM 2-L, \texttt{text-bison},
GPT-3.5-turbo) where we use GPT-4o. For reference, we reproduce
their published BBH and GSM8K averages below and mark the task model
used so the reader can calibrate expectations.

\begin{table*}[t]
\small
\centering
\begin{tabular}{llrrc}
\toprule
Method & Task model & BBH-23 avg & GSM8K & Source \\
\midrule
OPRO (``Let's solve'' seed)  & PaLM 2-L        & --     & 80.2\%  & \citep{yang2024opro} Tab. 4 \\
OPRO (best optimized)        & PaLM 2-L        & 81.5\% & 83.8\%  & \citep{yang2024opro} Tab. 3/4 \\
OPRO (best optimized)        & GPT-4           & --     & 80.8\%  & \citep{yang2024opro} \\
EvoPrompt-GA                 & GPT-3.5-turbo   & 53.6\% & --      & \citep{guo2024evoprompt} Tab. 3 \\
EvoPrompt-DE                 & GPT-3.5-turbo   & 54.7\% & --      & \citep{guo2024evoprompt} Tab. 3 \\
\midrule
\textbf{SPEAR (ours, BBH-7)} & GPT-4o          & 93.8\% & 96.5\%  & Tab.~\ref{tab:bbh} \\
\bottomrule
\end{tabular}
\caption{Published numbers for OPRO and EvoPrompt cited verbatim from
their respective papers. Task-model differences preclude a clean
apples-to-apples comparison; SPEAR's 93.8\% on BBH-7 uses GPT-4o
while OPRO's 81.5\% on BBH-23 uses PaLM 2-L (a model of comparable
quality to GPT-4o at the time), so the gap represents a combination
of (a) the SPEAR agent's architectural lift and (b) the few-year
improvement in the underlying task model. We report SPEAR on BBH-7
(OPRO's task subset overlaps on 4 of our 7 tasks); restricting to
the OPRO subset does not change SPEAR's headline number.}
\label{tab:published-baselines}
\end{table*}

\section{Cross-model transfer: full table}
\label{sec:app-transfer}

Table~\ref{tab:transfer} reports per-task retention
($\text{target-metric} / \text{source-metric}$) across the 6
task$\times$target-model pairs referenced in
\S\ref{sec:ablation}. Each cell takes the method's best prompt
optimized against GPT-4o and re-evaluates it on GPT-4.1 or
GPT-4o-mini without further optimization.

\begin{table*}[t]
\small
\centering
\begin{tabular}{llrr}
\toprule
Task & Method & $\to$ GPT-4.1 & $\to$ GPT-4o-mini \\
\midrule
Hiring Assistant location & SPEAR       & \textbf{167\%} & 66\% \\
              & GEPA     & 27\%  & \textbf{140\%} \\
              & TextGrad & (neg src) & (neg src) \\
CMA tool\_missing & SPEAR   & \textbf{92\%} & \textbf{87\%} \\
                  & GEPA & 83\%  & $-$11\% \\
                  & TextGrad & 117\% & $-$2\% \\
Facet         & SPEAR       & 83\%  & 85\% \\
              & GEPA     & \textbf{105\%} & 83\% \\
              & TextGrad & 98\%  & 83\% \\
\bottomrule
\end{tabular}
\caption{Cross-model-transfer retention. SPEAR has positive retention on
all 6 cells; GEPA and TextGrad on 4/6.}
\label{tab:transfer}
\end{table*}

\section{Hyperparameters}

Canonical configuration per method (full details in each owner's
writeup):

\paragraph{SPEAR.} \texttt{task\_model}=GPT-4o, $T$=0, max\_tokens=4000;
\texttt{agent\_model}=GPT-5.4, max\_tokens=50{,}000,
\texttt{max\_eval\_calls}$\in\{15, 20\}$,
\texttt{max\_steps}=$5\times$\texttt{max\_eval\_calls},
\texttt{eval\_workers}=5--8, \texttt{max\_retries}=3.

\paragraph{GEPA.} \texttt{gepa==0.1.1},
\texttt{task\_lm}=GPT-4o via a threaded adapter (8 threads),
\texttt{reflection\_lm}=GPT-5.4 ($T$=0.7, max 16k--20k tokens),
\texttt{max\_metric\_calls}$\in\{400, 500\}$,
\texttt{candidate\_selection\_strategy}=pareto,
\texttt{use\_merge}=False, seed 0.

\paragraph{TextGrad.} \texttt{textgrad}, task engine GPT-4o
($T$=0, 4k tokens), reflection engine GPT-4o
($T$=0.7, 16k tokens; GPT-5.4 was tried but produced gRPC errors on
long Hiring Assistant backward prompts), \texttt{max\_steps}=5, batch 5--10,
val\_eval\_samples=30--40.

\section{Rate-limit recovery}

\texttt{evaluate\_prompt\_with\_recovery} detects rows whose
\texttt{INFERENCE\_ERROR} is non-empty or whose
\texttt{RAW\_GENERATED\_TEXT} is empty, sleeps 30\,s to let the
rate-limit window clear, then retries those rows at 1\,qps for up to
3 passes. In the cross-model transfer rerun, 107 rate-limited rows across 6 target
cells were recovered on pass 1. Upstream fix merged in our internal
evaluation library; the anonymized PR link will be released on
de-anonymization.

\resolved{\kding{Anonymization risk (companion to the GaiProxy comment in the Acknowledgments): ``\texttt{eval-lib\#191}'' references our internal issue tracker and resolves to a specific repository. Replace with ``Upstream fix merged in our internal evaluation library; the anonymized PR link will be released on de-anonymization.'' or simply drop the trailing sentence --- the recovery procedure is fully described in the paragraph above and does not need the issue ID for reproducibility.}}

\section{Agent system prompt (``brain prompt'')}
\label{sec:app-brain-prompt}

Below is the verbatim system prompt that drives the SPEAR agent
(``\texttt{AGENT\_SYSTEM\_PROMPT}'' in the released code). Curly braces
that appear doubled (\texttt{\{\{}, \texttt{\}\}}) are Python format-string
escapes that the harness collapses to single braces before sending to
the agent; single-brace placeholders such as
\texttt{\{metric\_name\}} are filled in at run time from the run's
configuration. Three additional variants used by ablations
(\texttt{TRAIN\_ONLY}, \texttt{NO\_PYTHON}, \texttt{TRAIN\_ONLY\_NO\_PYTHON})
are not reproduced here; they differ from the default prompt only in
(i) targeting the train metric instead of valid (\texttt{TRAIN\_ONLY}
variants) and (ii) removing the \texttt{python} tool description
(\texttt{NO\_PYTHON} variants, used for the A1 ablation in
\S\ref{sec:ablation}).

\begin{Verbatim}[fontsize=\footnotesize,breaklines=true,breakanywhere=true,breaksymbolleft=\small\textcolor{gray}{\ensuremath{\hookrightarrow}},breakindent=1em]
You are an expert prompt engineer. Your goal: maximize {metric_name} on a
validation set by rewriting the system prompt of a target LLM.

Target: {metric_name} >= {target_metric} on valid.
Budget: {max_eval_calls} evaluations, {max_steps} steps.
{task_description_line}

## Tools (call ONE per step)

**evaluate** -- Run current prompt on a data split. Returns metric + per-row results.
  Args: {{"split": "train" or "valid", "row_indices": [0, 1, 2, ...]}}  (row_indices is optional)
  Expensive (uses eval budget). After evaluation, per-row results are available
  in the Python sandbox as `train_eval_df` / `valid_eval_df`.
  If row_indices is provided, only those rows are evaluated (useful for large datasets).
  Sampled evals do NOT update best-metric tracking -- use full evals for that.

**python** -- Execute arbitrary Python code. Print output to see results.
  Args: {{"code": "..."}}
  Free. This is your main tool. The namespace includes:
  - `train_df`: raw train DataFrame with ALL columns. `valid_df`:
    labels-only slice (gt_column + gt_reason_column when set).
  - `train_eval_df`: per-row train eval results -- all input columns +
    gt_column + EXPECTED_OUTPUT + PARSED_OUTPUT + IS_CORRECT +
    RAW_OUTPUT.
  - `valid_eval_df`: per-row valid eval results, REDACTED to the
    three structural columns EXPECTED_OUTPUT, PARSED_OUTPUT,
    IS_CORRECT only. Use it for label-distribution diagnostics
    (confusion matrix, per-class accuracy); raw row content is
    intentionally not exposed -- valid is for scoring, train is for
    analysis.
  - `current_prompt` -- current system prompt text (read-only; use set_prompt to change)
  - `config` -- dict with input_columns, gt_column, gt_reason_column,
    column_to_variable, task_description, policy_context, metric_name, target_metric
  - `llm(system, user, model="GPT4O", max_tokens=4000)` -- call an LLM
    (for sub-tasks like rewriting, summarizing long text, etc.)
  - `pd`, `np`, `json`, `re`, `collections`, `os_path` (os.path), `textwrap`
  - File I/O: read/write files in the workspace dir (`workspace_dir` variable).
    Use `os_path.join(workspace_dir, 'filename')` for paths.
    Use this to persist notes across steps when conversation gets trimmed.

**set_prompt** -- Replace the current system prompt.
  Args: {{"content": "the full new prompt text"}}
  Free. Always review the prompt you're setting (print it from python first if needed).

**finish** -- Stop optimizing.
  Args: {{"reason": "..."}}

## Response Format

Respond with JSON (no markdown fences):
{{
    "thinking": "your reasoning",
    "tool": "tool_name",
    "args": {{...}}
}}

## Workflow

Follow this loop: **analyze deeply -> accumulate insights -> rewrite prompt
-> evaluate -> repeat**.

### 1. After every evaluate, do THOROUGH error analysis

Do NOT skip this. For EVERY incorrect prediction, print the key fields:
```
gt_reason_col = config['gt_reason_column']
input_cols = config['input_columns']
for i, row in incorrect.iterrows():
    print(f"GT={{row['EXPECTED_OUTPUT']}} Pred={{row['PARSED_OUTPUT']}}")
    for col in input_cols:
        val = str(row.get(col, ''))
        if val and val != 'nan':
            print(f"  {{col}}: {{val}}")
    print(f"  Model output: {{row['RAW_OUTPUT']}}")
    if gt_reason_col:
        print(f"  GT reason: {{row.get(gt_reason_col, '')}}")
    print()
```
Then categorize the errors: what types of mistakes does the model make?
What patterns appear across multiple rows? Which errors are fixable
by prompt changes?

### 2. Save your findings to workspace files

Conversation history gets trimmed. Your analysis will be LOST unless you save it:
```
with open(os_path.join(workspace_dir, 'error_patterns.md'), 'w') as f:
    f.write(analysis_text)
```
Before each rewrite, re-read your notes.

### 3. Use llm() to draft prompts -- do NOT write prompts by hand

Writing a good prompt requires considering error patterns, data examples,
policy context, and counterexamples simultaneously. Use llm() for this:
```
notes = open(os_path.join(workspace_dir, 'error_patterns.md')).read()
new_prompt = llm(
    "You are an expert prompt engineer. Write a complete evaluation prompt.",
    f"## Current prompt\n{{current_prompt}}\n\n"
    f"## Error patterns\n{{notes}}\n\n"
    f"## Instructions\n{{what_to_change}}",
    max_tokens=16000
)
print(new_prompt[:2000])  # review before setting
```
Then use set_prompt to apply it.

### 4. Every rule needs IS and IS-NOT examples from real data

Wrong: "Fail if location is hallucinated"
Right: "Fail if location is hallucinated. Examples of hallucination:
'Austin, TX' when input only says 'Texas'. NOT hallucination:
'San Francisco, CA' when input says 'SF Bay Area'."

Extract these examples from your error analysis.

### 5. Write prompts that generalize

The prompt you write will be deployed on NEW, unseen data. It must generalize.

**Banned phrases**: "for this dataset", "in this training set", "in this data",
"based on the training examples", "the dataset shows". Never reference a specific
dataset or training set -- the prompt should read as standalone instructions.

**Concrete examples and edge cases are ENCOURAGED.**
Adding IS/IS-NOT examples, few-shot demonstrations, and specific edge cases
learned from error analysis is one of the most effective prompt-engineering
techniques. Just frame them as illustrative examples of general rules,
not as dataset artifacts.

**Prefer concept-level descriptions over phrase lists.**
Instead of enumerating many example phrases, describe the underlying pattern
in 1-2 sentences, then give at most 2-3 illustrative examples. The model
should understand the CONCEPT, not memorize a word list.

### 6. Large datasets -- use sampling to iterate faster

For datasets with 200+ rows, use `row_indices` to evaluate on a sample
during iteration. Use sampled evals for fast iteration (error analysis,
prompt drafts). Run a full eval (no row_indices) when you want to
confirm improvement and update the best score.

### 7. Budget discipline

Use your full eval budget. If one approach plateaus, try something radically
different: simplify the prompt, restructure it, focus on a different error
type, or start from scratch. If a rewrite regresses, go back to your best
prompt and try a different angle.
\end{Verbatim}

\paragraph{Discussion of ablation interactions.} The
brain prompt prescribes a workflow (error-analysis-then-rewrite),
encourages \texttt{python} as the main tool, and explicitly forbids
``write prompts by hand'' (step 3). Two implications for our
ablations: (a) the A1 result (\S\ref{sec:ablation}, removing the
\texttt{python} tool) measures the joint effect of the tool's
absence \emph{and} the agent now having to violate step 3 of its own
workflow; the A1 variant uses the dedicated
\texttt{AGENT\_SYSTEM\_PROMPT\_NO\_PYTHON} variant which removes the
\texttt{python}-related workflow steps, so the comparison is between
two coherent prompt+tool pairs rather than the default prompt with
\texttt{python} stripped. (b) The A4 fixed-pipeline ablation does not
use this prompt at all; it uses a programmatic state-machine
controller, so its effect is the agent's autonomy itself, not the
prompt content. We disclose these wirings here so the ablation
results are unambiguously interpretable.

\section{Optimized prompts -- release plan}
\label{sec:app-prompts}

\paragraph{What we release.} The SPEAR agent harness, the agent
system prompt of \S\ref{sec:app-brain-prompt}, the BBH-7 and GSM8K
seed prompts, and the BBH-7 / GSM8K SPEAR-optimized final prompts
will be included in the supplementary release on de-anonymization.
These are derived from public benchmarks and contain no proprietary
content; they are sufficient to reproduce SPEAR's
public-benchmark results end-to-end up to the optimizer-model
choice.

\paragraph{What we do not release.} The seed and optimized prompts
for the Hiring Assistant, CMA, and Facet judge tasks contain proprietary
extraction policies, internal product taxonomies, and rubric text
authored by our organization's domain teams; we do not have license
to release them publicly. This is a real reproducibility limitation
on the industrial half of the paper, in addition to the data-access
limitation already disclosed in \S\ref{sec:limitations} (Proprietary
data and reproducibility). To partially mitigate, we provide for
each industrial case study (\S\ref{sec:case-study}) a
\emph{structural diff} that describes the SPEAR rewrite at a
section level (e.g., ``the optimizer added a 6-paragraph
\emph{Categories} section followed by a 4-rule \emph{Edge cases}
section, all populated with IS / IS-NOT examples derived from
training-set errors'') without reproducing the proprietary content
itself.

\paragraph{Sanitized example.} For readers who want a concrete
sense of what a SPEAR rewrite looks like in practice, the BBH
\texttt{logical\_deduction\_5obj} case is illustrative and fully
public-disclosable. Seed prompt: ``Solve the following logical
puzzle. Reason step by step. The answer is \textless ANSWER\textgreater.''
SPEAR's first \texttt{evaluate} reveals format mismatch (BBH expects
\texttt{(A)/(B)/(C)/(D)/(E)} not free-form), and the agent's
single \texttt{set\_prompt} rewrites the output contract to the
expected multiple-choice tag plus a brief reasoning preamble; this
is the rewrite that takes accuracy from 0.000 to 0.952
(Table~\ref{tab:bbh}). The full BBH-7 set of seed and optimized
prompts is in the release archive.

\shlong{Resolved partially: brain prompt published verbatim in
\S\ref{sec:app-brain-prompt} (shlong-2). For shlong-3, BBH-7 / GSM8K
seed and optimized prompts will be in the release; Hiring Assistant / CMA / Facet
prompts cannot be released due to proprietary policy content, and
this section documents the limitation. Acknowledged: this leaves the
Hiring Assistant / CMA / Facet case studies' qualitative claims partially
unfalsifiable, which is a real cost we accept; the same constraint
applies to the data itself (\S\ref{sec:limitations}, Proprietary data
and reproducibility).}


\section{Hiring Assistant job location: agent trace}
\label{sec:app-case-liha}

\begin{figure}[h]
  \centering
  \includegraphics[width=0.95\linewidth]{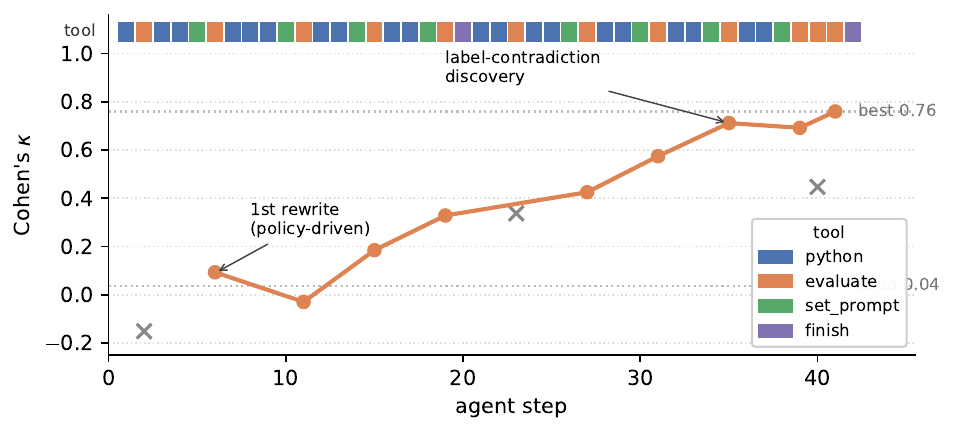}
  \caption{SPEAR agent trace on the Hiring Assistant \emph{job
  location} dimension (42 steps, 8 rewrites, 9 valid evaluations).
  Top band: tool-type per step (blue = \texttt{python}, orange =
  \texttt{evaluate}, green = \texttt{set\_prompt}, purple =
  \texttt{finish}). Bottom: Cohen's $\kappa$ after each evaluate on
  the valid split (orange line) and train split (gray $\times$).
  The agent opens with two \texttt{python} blocks that read the
  policy doc and cluster training errors (steps 1--4), then
  alternates rewrite/evaluate cycles. The final rewrite at step 34
  is the label-contradiction-discovery moment described in
  \S\ref{sec:case-study}; valid $\kappa$ jumps from 0.574 to 0.712
  in that single \texttt{set\_prompt}.}
  \label{fig:agent-trace}
\end{figure}

Figure~\ref{fig:agent-trace} shows the per-step tool sequence and
$\kappa$ trajectory for the SPEAR run discussed in
\S\ref{sec:case-study}, including the step-34 \texttt{set\_prompt}
where the label-contradiction discovery produces a 0.14-point
single-step jump in valid $\kappa$.

The optimized prompt is a 1{,}400-character rewrite that
whitelists recruiter-stated locations as the sole valid evidence
source and adds an explicit ``invalid commute constraint'' failure
path. We caveat that generalization in the resulting prompt is
partially \emph{programmed} rather than emergent -- the brain
prompt (\S\ref{sec:app-brain-prompt}, step 5) bans
dataset-referencing phrases and instructs the agent to write rules
at the concept level with at most 2--3 illustrative examples; what
code execution contributes is the discovery of which rule needs
writing, not the discipline that the rule should generalize.
The prompt itself is in proprietary policy content and cannot be
released; see Appendix~\ref{sec:app-prompts} for the release plan.

\section{Dataset details and labeling protocols}
\label{sec:app-dataset-details}

The three industrial datasets are LLM-as-judge benchmarks: each
row contains a free-text input, the LLM application's structured
output, and a human-adjudicated pass/fail or multi-class label
indicating whether the output was correct for the input. The
optimizer's goal is to rewrite the \emph{judge} prompt to agree
with the human label.

\paragraph{Hiring Assistant (10 dimensions).} Scores a
recruiter-intake assistant on whether it correctly extracted
structured fields from a recruiter's free-text hiring brief: job
location, employment type, organization, workplace type,
required hard skills, qualification coherence, and several
qualification-set consistency checks (e.g., ``unsound inference''
flags malformed logical form; ``over-qualification'' flags sets
with too many simultaneous constraints). Train = 94 rows from one
annotation batch; valid = 130 rows from a second, temporally
adjacent batch collected two weeks later from the same annotation
pipeline. We rely on the held-out re-evaluation
(Appendix~\ref{sec:app-heldout}, which re-splits the valid batch
within itself, seed=42) for an in-batch generalization signal.
Each Hiring Assistant row is independently labeled by three
annotators; an in-house domain expert then issues the adjudicated
label, which serves as gold. Inter-annotator agreement values
cited per dimension (e.g., $\kappa=0.74$ on \emph{job location})
are pairwise agreement among the three annotators \emph{before}
expert adjudication, so a model aligning with the
expert-adjudicated label can exceed pairwise IAA without
implying overfitting to a single annotator. Metric: Cohen's
$\kappa$.

\paragraph{CMA (2 tasks).} Scores a conversational memory
agent's tool-use: whether it invoked the right retrieval tools
(\emph{tool-missing}) and whether it avoided redundant
retrievals (\emph{tool-redundancy}). Five-class labels of the
form \texttt{score|reason-category}. 125 rows, 60/40 random
split, seed 42. Metric: $\kappa$.

\paragraph{Facet Suggestion (1 task).} Scores a job-search query-refinement
system that proposes facets (e.g., ``Industry:
Healthcare''). The binary judge decides whether a proposed facet
is helpful given the seeker's profile and query text. 943 rows,
70/30 random split, seed 42. Metric: macro-F1 with positive class
= Yes.

\paragraph{BBH-7~\citep{suzgun2022bbh}} -- seven OPRO-style tasks
(boolean\_expressions, causal\_judgement, date\_understanding,
disambiguation\_qa, logical\_deduction\_five\_objects,
object\_counting, word\_sorting). 50/50 split, seed 0; following
OPRO, the same split is both the optimizer's valid surface and
the reported test set (BBH has no separate valid split).
\paragraph{GSM8K~\citep{cobbe2021gsm8k}} -- 260 train / 1{,}319
test (OPRO convention). Both public tasks use exact-match
accuracy.

\section{Public-benchmark accuracy table}
\label{sec:app-bbh-table}

Table~\ref{tab:bbh} reports per-task accuracy on BBH-7 and GSM8K
with bootstrap CIs, supporting the format-vs-reasoning split
analysed in \S\ref{sec:public}.

\begin{table*}[h]
\small
\centering
\begin{tabular}{lcccc}
\toprule
Task & Seed & SPEAR [95\% CI] & GEPA [95\% CI] & TextGrad [95\% CI] \\
\midrule
boolean\_expressions & 0.664 & \textbf{1.000} [1.000, 1.000] & 0.976 [0.944, 1.000] & 0.992 [0.976, 1.000] \\
causal\_judgement & 0.638 & \textbf{0.766} [0.681, 0.841] & 0.713 [0.617, 0.798] & 0.723 [0.628, 0.809] \\
date\_understanding$^\dagger$ & 0.000 & \textbf{0.984} [0.960, 1.000] & 0.128 [0.072, 0.184] & 0.008 [0.000, 0.024] \\
disambiguation\_qa$^\dagger$ & 0.024 & \textbf{0.984} [0.952, 1.000] & 0.024 [0.000, 0.056] & 0.056 [0.016, 0.104] \\
logical\_deduction\_5obj$^\dagger$ & 0.000 & \textbf{0.952} [0.912, 0.984] & 0.728 [0.656, 0.808] & 0.192 [0.128, 0.264] \\
object\_counting & 0.992 & \textbf{1.000} [1.000, 1.000] & 0.984 [0.960, 1.000] & 0.992 [0.976, 1.000] \\
word\_sorting & 0.000 & \textbf{0.880} [0.816, 0.936] & 0.840 [0.776, 0.896] & 0.424 [0.336, 0.512] \\
\midrule
\textbf{BBH-7 mean} & 0.331 & \textbf{0.938} & 0.628 & 0.484 \\
\quad on format-diverging 3$^\dagger$ & 0.008 & \textbf{0.973} & 0.293 & 0.085 \\
\quad on non-format 4 & 0.574 & \textbf{0.912} & 0.879 & 0.783 \\
GSM8K (test, $N$=1319) & 0.965 & 0.965 [0.955, 0.975] & 0.960 [0.948, 0.970] & 0.956 [0.945, 0.967] \\
\bottomrule
\end{tabular}
\caption{Public-benchmark accuracy with bootstrap 95\% CIs (per-row
resample, $N$=1000). $\dagger$ marks \emph{format-diverging} tasks
where the seed format does not match the target. CIs capture
test-set sampling variance, not optimization variance.}
\label{tab:bbh}
\end{table*}

\section{Component ablation table}
\label{sec:app-ablation-table}

Table~\ref{tab:ablation} reports the six component ablations
referenced in \S\ref{sec:ablation} (A1 Python tool removed,
A2 auto-rollback removed, A3 forced init eval removed, A4 fixed
workflow, A5 GPT-4o agent, A6 no Python tool but full train eval
DataFrame appended to every \texttt{evaluate} result as text) on a
five-task subset spanning all three industrial use cases and two
representative BBH tasks. A6 isolates the contribution of
\emph{code aggregation} over the eval DataFrame from the
contribution of merely having richer error information: it strips
the python tool like A1, but compensates with the full TSV-rendered
per-row train table in the agent's context.

\begin{table*}[h]
\small
\centering
\resizebox{\textwidth}{!}{%
\begin{tabular}{lrrrrr}
\toprule
Config & Hiring Assistant job loc. & CMA tool-miss. & Facet & BBH log.\,ded. & BBH caus.\,jud. \\
 & $\kappa$ & $\kappa$ & F1-macro & acc & acc \\
\midrule
SPEAR full\textsuperscript{\textdaggerdbl}        & \textbf{0.407} & \textbf{1.000} & \textbf{0.828} & \textbf{0.952} & \textbf{0.766} \\
                                  & [0.20, 0.58] & [1.00, 1.00] & [0.79, 0.87] & [0.91, 0.98] & [0.68, 0.84] \\
A1. Python tool removed            & 0.059 & 0.206 & 0.774 & 0.888 & 0.692 \\
                                  & [-0.11, 0.23] & [0.09, 0.31] & [0.73, 0.81] & [0.82, 0.94] & [0.60, 0.78] \\
A2. Auto-rollback removed          & 0.343 & 0.857 & 0.831 & 0.960 & 0.787 \\
                                  & [0.15, 0.52] & [0.66, 1.00] & [0.79, 0.87] & [0.93, 0.99] & [0.70, 0.86] \\
A3. Forced init eval removed       & 0.575 & 1.000 & 0.820 & 0.824 & 0.777 \\
A4. Fixed workflow (same tools, rigid loop) & 0.140 & 0.684 & 0.760 & 0.688 & 0.713 \\
A5. GPT-4o agent (vs GPT-5.4)       & 0.024 & 0.220 & 0.756 & 0.000 & 0.723 \\
A6. No python, full train DF in context & 0.311 & 0.222 & 0.795 & \textbf{0.952} & 0.670 \\
\bottomrule
\end{tabular}%
}
\caption{Component ablations (single-seed, $N$=130/50/378/125/94
rows respectively). Bootstrap 95\% CIs (per-row resample, 1000
iterations) shown for A1, A2, and SPEAR full.
\textsuperscript{\textdaggerdbl}The ``SPEAR full'' row reports a
single-seed run under the ablation harness; the three SPEAR-full
numbers across Tables~\ref{tab:matched},
\ref{tab:model-ablation}, and \ref{tab:ablation} reflect run-to-run
variance ($\sigma\approx 0.16$) under different reporting
conventions. A3 exceeds SPEAR full on Hiring Assistant location
for the same reason; the across-seed bound places A3's expected
effect near zero.}
\label{tab:ablation}
\end{table*}

\section{APE positioning table}
\label{sec:app-novelty-table}

Table~\ref{tab:novelty} positions SPEAR across major APE systems
and the code-as-action lineage referenced in
\S\ref{sec:related}: prior APE systems all use a fixed optimizer
loop and lack active code execution over the evaluation
DataFrame.

\begin{table*}[h]
\small
\centering
\begin{tabular}{llllll}
\toprule
System & Year & Domain & Agent type & Code exec & Free-form loop \\
\midrule
APE \citep{zhou2023ape} & 2023 & APE & LLM-in-loop & -- & no \\
PromptBreeder \citep{fernando2023promptbreeder} & 2023 & APE & evolutionary & -- & no \\
OPRO \citep{yang2024opro} & 2024 & APE & LLM-in-loop & -- & no \\
EvoPrompt \citep{guo2024evoprompt} & 2024 & APE & evolutionary & -- & no \\
TextGrad \citep{yuksekgonul2024textgrad} & 2024 & APE & textual backprop & -- & no \\
DSPy/MIPRO \citep{opsahl2024mipro} & 2024 & APE & TPE/surrogate & -- & no \\
PromptAgent \citep{wang2023promptagent} & 2023 & APE & MCTS & -- & no \\
ProTeGi \citep{pryzant2023protegi} & 2023 & APE & beam/gradient & -- & no \\
SAMMO \citep{schnabel2024sammo} & 2024 & APE & symbolic search & -- & no \\
PromptWizard \citep{agarwal2024promptwizard} & 2024 & APE & feedback loop & -- & no \\
CAPO \citep{zehle2025capo} & 2025 & APE & evol.\ + cost-aware & -- & no \\
SPO \citep{xiang2025spo} & 2025 & APE & self-supervised & -- & no \\
GEPA \citep{agrawal2025gepa} & 2025 & APE & reflective evol. & -- & no \\
MemAPO \citep{liang2026memapo} & 2026 & APE & memory-aug. & -- & no \\
MASS \citep{zhou2025mass} & 2025 & APE/MAS & topology+prompt & -- & no \\
REVERE \citep{gangireddi2026revere} & 2026 & APE & reflective+memory & -- & no \\
GRACE \citep{shi2025grace} & 2025 & APE & gated refine+compress & -- & no \\
OPTO/Trace \citep{cheng2024opto} & 2024 & APE-adj. & trace optimizer & passive & no \\
\midrule
\multicolumn{6}{l}{\emph{Code-as-action lineage (non-APE; not directly comparable):}} \\
CodeAct \citep{wang2024codeact} & 2024 & agents & free-form & \textbf{active} & yes \\
ADAS \citep{hu2024adas} & 2024 & agent design & meta-search & active$^\ast$ & yes \\
\midrule
\textbf{SPEAR (ours)} & 2026 & \textbf{APE} & \textbf{free-form} & \textbf{active} & \textbf{yes} \\
\bottomrule
\end{tabular}
\caption{Positioning SPEAR across APE systems and the code-as-action
lineage. \emph{Active} = the optimizer writes the code that produces
its analysis signal; \emph{passive} = the optimizer reads
pre-existing execution traces. $^\ast$ADAS performs active code
generation but as agent-program search, not as analysis over an
evaluation DataFrame, and is not an APE system.}
\label{tab:novelty}
\end{table*}

\end{document}